\documentclass[10pt,journal,compsoc]{IEEEtran}

\ifCLASSOPTIONcompsoc
  \usepackage[nocompress]{cite}
\else
  \usepackage{cite}
\fi

\usepackage{nicefrac}
\usepackage{multirow}
\usepackage{dblfloatfix}

\usepackage{graphicx,xfp}
\usepackage{multicol}
\usepackage{multirow}
\usepackage{color}
\usepackage{xcolor}
\usepackage{booktabs}
\usepackage{graphicx}
\usepackage{nicefrac}
\usepackage{amsmath}
\usepackage{array}
\newcolumntype{P}[1]{>{\centering\arraybackslash}p{#1}}

\usepackage{tikz}
\usetikzlibrary{positioning}
\usepackage{pgfplots}

\usepackage[multiple]{footmisc}
\usepackage{tikzscale}
\usepackage{xspace}
\usepgfplotslibrary{groupplots}
\pgfplotsset{compat=1.16}

\usepackage{hyperref}
\usepackage[switch]{lineno}
\usepackage{lipsum}

\DeclareRobustCommand*\cal{\relax\mathcal}

\usepackage{caption}
\usepackage{subcaption}

\usepackage{amssymb}
\usepackage{amsmath}
\usepackage{flexisym}

\usepackage{amsmath,amssymb}
\usepackage{amsthm}
\usepackage{algorithm}
\usepackage[noend]{algpseudocode}
\usepackage{hyperref}
\usepackage{verbatim}
\usepackage{amssymb}
\usepackage{tablefootnote}

\def\beq{\begin{equation}}\def\eeq{\end{equation}}

\def\l{\left(}
\def\r{\right)}

\newcommand{\Q}{\mathcal{Q}}

\renewcommand{\vec}[1]{\mathbf{#1}}

\newtheorem{theorem}{Theorem}[section]
\newtheorem{lemma}[theorem]{Lemma}

\newtheorem{remark}[theorem]{Remark}

\newtheorem{problem}[theorem]{Problem}

\newtheorem{fact}[theorem]{Fact}
\newtheorem{corollary}[theorem]{Corollary}

\begin{document}
%
\title{Efficient Approximate Kernel Based Spike Sequence Classification}

\author{
Sarwan Ali, Bikram Sahoo, Muhammad Asad Khan , Alexander Zelikovsky, Imdad Ullah Khan*, Murray Patterson*%
\IEEEcompsocitemizethanks{\IEEEcompsocthanksitem S. Ali, B. Sahoo, A.Zelikovskiy, and M. Patterson are with the Department
of Computer Science, Georgia State University, Atlanta,
GA, USA, 30303.\protect\\
E-mail: \{sali85,bsahoo1\}@student.gsu.edu, \\
{\ alexz,mpatterson30\}@gsu.edu}
\IEEEcompsocthanksitem M. A. Khan is with Department of Telecommunication Engineering, Hazara University, Mansehra, Pakistan.
\protect\\
E-mail: asadkhan@hu.edu.pk
\IEEEcompsocthanksitem I. Khan is with Department of Computer Science, Lahore University of Management Sciences, Lahore, Pakistan.
\protect\\
E-mail: imdad.khan@lums.edu.pk
}%
\thanks{* Joint Last/corresponding Authors}
\thanks{Manuscript received February 25, 2022.}
}


\markboth{Transactions on Computational Biology and Bioinformatics (TCBB)}%
{Shell \MakeLowercase{\textit{et al.}}: Bare Demo of IEEEtran.cls for Computer Society Journals}

\IEEEtitleabstractindextext{%
\begin{abstract}
Machine learning (ML) models, such as SVM, for tasks like classification and clustering of sequences, require a definition of distance/similarity between pairs of sequences. Several methods have been proposed to compute the similarity between sequences, such as the exact approach that counts the number of matches between $k$-mers (sub-sequences of length $k$) and an approximate approach that estimates pairwise similarity scores. Although exact methods yield better classification performance, they pose high computational costs, limiting their applicability to a small number of sequences. The approximate algorithms are proven to be more scalable and perform comparably to (sometimes better than) the exact methods --- they are designed in a ``general'' way to deal with different types of sequences (e.g., music, protein, etc.). Although general applicability is a desired property of an algorithm, it is not the case in all scenarios. For example, in the current COVID-19 (coronavirus) pandemic, there is a need for an approach that can deal specifically with the coronavirus. To this end, we propose a series of ways to improve the performance of the approximate kernel (using minimizers and information gain) in order to enhance its predictive performance pm coronavirus sequences. More specifically, we improve the quality of the approximate kernel using domain knowledge (computed using information gain) and efficient preprocessing (using minimizers computation) to classify coronavirus spike protein sequences corresponding to different variants (e.g., Alpha, Beta, Gamma). 
We report results using different classification and clustering algorithms and evaluate their performance using multiple evaluation metrics.
Using two datasets, we show that our proposed method helps improve the kernel's performance compared to the baseline and state-of-the-art approaches in the healthcare domain. 
\end{abstract}

}

\maketitle

\IEEEdisplaynontitleabstractindextext
\IEEEpeerreviewmaketitle


\IEEEraisesectionheading{\section{Introduction}\label{sec:introduction}}

\IEEEPARstart{T}{he} COVID-19 global pandemic is unique compared to past pandemics because it occurs in a time of worldwide travel like never before and the widespread availability of high-throughput sequencing. Before this pandemic, sequences for a given virus were gathered in the order of hundreds, maybe a few thousand, but the current number of sequences of the SARS-CoV-2 virus (which causes the COVID-19 disease) is orders of magnitude beyond this. This amount is so high that methods such as phylogenetic tree building~\cite{hadfield2018a}, which have traditionally been used for studying the diversity, dynamics, and evolution of viruses, are no longer appropriate in this situation because they do not scale. Some researchers have hence turned to alternatives such as clustering and classification to tackle this ``big data'' problem~\cite{melnyk2021alpha,ali2021k,ali2021spike2vec}.

\begin{figure}[!ht]
  \centering
  \includegraphics[scale=0.25,page=1]{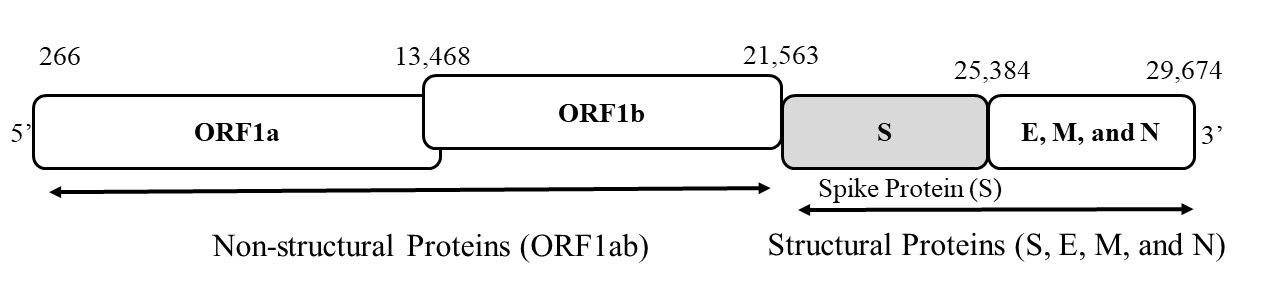}
  \caption{The SARS-CoV-2 genome is around 29--30kb, encoding structural and non-structural proteins.  ORF1ab encodes the non-structural proteins, while the four structural proteins spike, envelope, membrane, and nucleocapsid are encoded by their respective genes.  The spike protein has roughly 1300 amino acids.}
  \label{fig_spike_seq_example}
\end{figure}

The SARS-CoV-2 is a type of coronavirus, so-called because of its notable spikes, which resemble ``crowns''. These spikes serve as the mechanism for the virus to fuse to the host cell membrane. Coronaviruses such as SARS-CoV-2 cause a wide variety of respiratory diseases in an array of different hosts. Changes in these spikes (in the form of spike protein mutations) allow coronaviruses to adapt to different hosts and evolve into new and more transmissible variants. See Figure~\ref{fig_spike_seq_example} for an illustration of the SARS-CoV-2 genomic structure, including the region (the spike region) that codes the spike protein. It is hence important to use this source of information to identify different host specificity~\cite{ali2022pwm2vec} and variants~\cite{ali2021spike2vec}. This motivates approaches for classifying coronavirus spike sequences to better understand the dynamics of the different variants in terms of this information.

Most classification approaches leverage the powerful tools of machine learning (ML) --- methods such as support vector machine (SVM) or random forest (RF), which have highly optimized implementations. One problem researchers face while utilizing ML models' power is to convert the character-based sequences into fixed-length numerical vectors so that the machine can understand them. One popular way to deal with this problem is by using $k$-mers-based approach (a substring of length $k$).
Several alignment-based~\cite{kuzmin2020machine,ali2022pwm2vec} and alignment-free~\cite{ali2021spike2vec,ali2021effective} have been proposed recently for ML tasks such as classification and clustering. Most of these methods involved computing $k$-mers from the spike sequences and then computing the $k$-mers count to get a frequency vector (more detail about $k$-mers is given in Section~\ref{sec_omk}). 
Since these methods are based on feature engineering, they may not fully utilize ML models' power as loss of information may occur while performing the feature engineering process.
One problem with the $k$-mers-based methods is that the number of $k$-mers in a given sequence could be large (and there could be a large number of similar/redundant $k$-mers). Therefore, there is a computational overhead for processing these redundant $k$-mers.
One way to reduce that overhead is by using \emph{Minimizers}~\cite{roberts_2004_minimizer}.
Minimizers are a type of lightweight ``signature'' of a sequence that is used primarily in the context of \emph{de novo} assembly and read mapping. In this paper, we are using minimizers (as a pre-processing step) as a way to remove some of the amino acids from the sequences and show that this eventually improves the predictive performance of the overall algorithm for classification and clustering.

A popular domain of research for sequence classification that has shown success in the past is using kernel-based method~\cite{Kuksa_SequenceKernel,farhan2017efficient}. These methods compute the exact/approximate distance between pairs of sequences based on the matches and mismatches between the $k$-mers (substrings of length $k$) of the sequences. In~\cite{ali2021k}, the authors use an approximate kernel proposed in~\cite{farhan2017efficient} to compute the distance between pairs of sequences. These kernels are then used to perform classification using kernel-based classifiers such as SVM and non-kernel-based classifiers (using kernel PCA) such as decision trees. In this work, we devise a kernel that is computationally more efficient, has the kernel's theoretical properties and yields excellent predictive accuracy for both clustering and classification. We then use this kernel in comparison with many baselines (including that of~\cite{ali2021k}) and show that it outperforms all baselines in terms of predictive performance and runtime.

Our contributions in this paper are the following:
\begin{enumerate}
  \item We propose a method based on minimizers, information gain, and approximate kernel to perform classification and clustering on the COVID-19 spike sequences
    \item Unlike in~\cite{ali2021k}, we use more variants to show the performance of kernel matrix with a higher number of classes.
    \item We show that spike sequences alone can be used to classify different COVID-19 variants efficiently.
    \item We show that our method could work for classification and clustering tasks.
    \item Using the domain knowledge (information gain), we show that the classification and clustering performance could be improved compared to the baselines.
    \item We prove that the proposed kernel is positive semidefinite.
\end{enumerate}

The rest of the paper is organized as follows:
Section~\ref{sec_related_work} contains related work. Our proposed
model is given in Section~\ref{section:algo}, Section~\ref{sec_omk}, Section~\ref{sec_igk}, and Section~\ref{sec_omkig}.
Section~\ref{sec_exp_setup} contains the experimental setup. Results
are given in Section~\ref{sec_results}. Finally, we conclude the paper
in Section~\ref{sec_conclusion}.

\section{Related Work}
\label{sec_related_work}
In bioinformatics, sequence homology (shared ancestry) detection between a pair of proteins and prediction of disease transmission using the Phylogeny-based method~\cite{Dhar2020TNet} are essential tasks. The use of $k$-mers counts for phylogenetic applications was first explored in~\cite{Blaisdell1986AMeasureOfSimilarity}, which proposed constructing accurate phylogenetic trees from several coding and non-coding nDNA sequences. In bioinformatics, sequence classification is a widely studied problem~\cite{Krishnan2021PredictingVaccineHesitancy}. Some classification methods are alignment-free (considered computationally less expensive), while others rely on sequences' alignment (comparatively more computationally expensive).

Converting input data into fixed-length numerical vectors for applying different machine learning algorithms such as classification and clustering is a common practice across numerous fields like smart grid~\cite{ali2019short,ali2019short_AMI}, graph analytics~\cite{ali2021predicting,AHMAD2020Combinatorial,Ahmad2016AusDM,Tariq2017Scalable,ahmad2017spectral}, electromyography~\cite{ullah2020effect}, clinical data analysis~\cite{ali2021efficient}, network security~\cite{ali2020detecting}, and text classification~\cite{Shakeel2020LanguageIndependent}. Authors in~\cite{ali2022pwm2vec} use the position weight matrix-based approach to compute feature embeddings for spike sequences. Although their approach shows promising results, one drawback of their method is that it only works for aligned data. A $k$-mers-based approach for classification of SARS-CoV-2 sequences is proposed in~\cite{ali2021spike2vec}. Several methods to perform clustering on the spike sequence data have also been proposed recently~\cite{ali2021effective,tayebi2021robust}.

Another domain for classifying sequences is by using kernel matrix (gram matrix). In this method, a kernel matrix is computed that represents the similarity between pairs of sequences~\cite{farhan2017efficient, Kuksa_SequenceKernel}. This matrix is used as input to kernel-based classifiers like Support Vector Machines (SVM). Recently, authors in~\cite{ali2021k} proposed a kernel-based approach for spike sequence classification. Although their method shows promising results on classification, it is not clear if the proposed method can be generalized to more variants and for other bioinformatics tasks such as clustering.

One issue with sequence similarity measures involving dynamic programming such as Smith-Waterman~\cite{smith1981identification}, is that they have quadratic ($O(n^2)$) runtime and space complexity.  Even at 7000 sequences, performing such an operation for all ($49$ million) pairs would be infeasible.  In fact, the approximate kernel method we propose here is precisely for this reason, it allows for pairwise comparisons to be much faster ($O(k^2 nlog \ n)$, where $k$ is a constant and $n$ refers to the length of larger sequence between a pair of sequences).  On the other hand, using common sequence similarity measures such as Hamming distance (which takes linear time) maybe too simple: not necessarily reflecting the underlying (e.g., physical) similarity between sequence fragments~\cite{Kuksa_SequenceKernel}. For example, in protein sequence analysis, different pairs of symbols (amino acids) induce
different similarity levels, a consequence of particular physical or chemical properties. Hence, treating all differences the same (similar weight) may not be a good option. Moreover, Hamming distance treats sequences as vectors, which means that the sequences must have same length, which is not always the case. As compared to our alignment-free kernel-based method, Hamming distance may not be applicable in real world because of its dependence on sequence alignment, which itself is an expensive operation.

\section{Algorithm for Kernel Computation} \label{section:algo}
In this section, we formulate the problem, describe our algorithm and analyze its runtime and quality.

\textbf{$k$-spectrum and $k,m$-mismatch kernel:} Given a sequence $X$ over alphabet $\Sigma$, the $k,m$-mismatch spectrum of $X$ is a $|\Sigma|^k$-dimensional vector, $\Phi_{k,m}(X)$ of number of times each possible $k$-mer occurs in $X$ with at most $m$ mismatches. Formally,
\begin{equation}
	\label{Eq:MismatchSpectrum}
	\Phi_{k,m}(X) =  \left( \Phi_{k,m}(X)[\gamma]\right)_{\gamma \in \Sigma^k} 
	=     \left( \sum_{\alpha \in X} I_m(\alpha,\gamma)\right)_{\gamma \in \Sigma^k},
\end{equation}
where $I_m(\alpha,\gamma)=1$, if $\alpha$ belongs to the set of $k$-mers that differ from $\gamma$ by at most $m$ mismatches, i.e. the Hamming distance between $\alpha$ and $\gamma$,  $d(\alpha,\gamma)\leq m$. Note that for $m=0$, it is known as \textit{$k$-spectrum} of $X$. The {\em $k,m$-mismatch} kernel value for two sequences $X$ and $Y$ (the mismatch spectrum similarity score) \cite{leslie2002mismatch} is defined as: 

\begin{equation} \label{Eq:MK}
\centering
\begin{aligned}
	K(X,Y|k,m) =  \langle \Phi_{k,m}(X),\Phi_{k,m}(Y)\rangle
\end{aligned}
\end{equation}

\begin{equation*} 
\centering
\begin{aligned}
	=  \sum_{\gamma \in \Sigma^k} \Phi_{k,m}(X)[\gamma] \Phi_{k,m}(Y)[\gamma]\notag
	\\
	=  \sum_{\gamma \in \Sigma^k} \sum_{\alpha \in X} I_m(\alpha,\gamma) \sum_{\beta \in Y} I_m(\beta,\gamma)
	\\
	=  \sum_{\alpha \in X} \sum_{\beta \in Y} \sum_{\gamma \in \Sigma^k} I_m(\alpha,\gamma) I_m(\beta,\gamma).
\end{aligned}
\end{equation*}

For a $k$-mer $\alpha$, let $N_{k,m}(\alpha)=\{\gamma\in \Sigma^k: d(\alpha,\gamma)\leq m\}$ be the {\em $m$-mutational neighborhood} of $\alpha$. Then for a pair of sequences $X$ and $Y$, the $k,m$-mismatch kernel given in eq (\ref{Eq:MK}) can be equivalently computed as follows \cite{kuksa2009scalable}: 
\beq \label{Eq:MK1}
\begin{aligned}
	K(X,Y|k,m) =&\sum_{\alpha \in X} \sum_{\beta \in Y} \sum_{\gamma \in \Sigma^k} I_m(\alpha,\gamma) I_m(\beta,\gamma)
\end{aligned}
\eeq

\begin{equation*} 
\begin{aligned}
	= \sum_{\alpha \in X} \sum_{\beta \in Y} |N_{k,m}(\alpha) \cap N_{k,m}(\beta)| 
	\\
	=  \sum_{\alpha \in X} \sum_{\beta \in Y} {\mathfrak{I}}_m(\alpha,\beta),
\end{aligned}
\end{equation*}

where ${\mathfrak{I}}_m(\alpha,\beta) = |N_{k,m}(\alpha) \cap N_{k,m}(\beta)|$ is the size of intersection of $m$-mutational neighborhoods of $\alpha$ and $\beta$. We use the following two facts.

\begin{fact}\label{independence}
	${\mathfrak{I}}_m(\alpha,\beta)$, the size of the intersection of $m$-mismatch neighborhoods of $\alpha$ and $\beta$, is a function of $k$, $m$, $|\Sigma|$ and $d(\alpha,\beta)$ and is independent of the actual $k$-mers $\alpha$ and $\beta$ or the actual positions where they differ.
\end{fact} 
\begin{fact}\label{min2mk}
	If $d(\alpha,\beta) > 2m$, then ${\mathfrak{I}}_m(\alpha,\beta) = 0$. 
\end{fact}
In view of the above two facts, we can rewrite the kernel value (\ref{Eq:MK1}) as 
\beq \label{Eq:MK2} K(X,Y|k,m) =  \sum_{\alpha \in X} \sum_{\beta \in Y} {\mathfrak{I}}_m(\alpha,\beta) = \sum_{i=0}^{\min\{2m,k\}} M_i(X,Y)\cdot {\cal I}_i,\eeq where ${\cal I}_i={\mathfrak{I}}_m(\alpha,\beta)$ when $d(\alpha,\beta)=i$ and $M_i(X,Y)$ is the number of pairs of $k$-mers $(\alpha,\beta) $ such that $d(\alpha,\beta)=i$, where $\alpha\in X$ and $\beta\in Y$. Note that bounds on the last summation follow from Fact \ref{min2mk} and the fact that the Hamming distance between two $k$-mers is at most $k$. Hence the problem of kernel evaluation is reduced to computing $M_i(X,Y)$'s and evaluating ${\cal I}_i$'s.

A closed form formula to evaluate the size of intersection of mismatch neighborhoods of two $k$-mers at distance $i$ is derived in \cite{farhan2017efficient}. Let $N_{k,m}(\alpha,\beta)$ be the intersection of $m$-mismatch neighborhoods of $\alpha$ and $\beta$ i.e.
\begin{equation*}
    N_{k,m}(\alpha,\beta) = N_{k,m}(\alpha) \cap N_{k,m}(\beta)
\end{equation*}
As defined earlier $|N_{k,m}(\alpha,\beta)| = {\mathfrak{I}}_m(\alpha,\beta)$. Let $N_q(\alpha) = \{\gamma \in \Sigma^k : d(\alpha,\gamma) = q\}$ be the set of $k$-mers that differ with $\alpha$ in exactly $q$ indices. Note that $N_q(\alpha) \cap N_r(\alpha) = \emptyset$ for all $q\neq r$. 
Using this and defining $n^{qr}(\alpha,\beta)=|N_q(\alpha) \cap N_r(\beta)|$,

\begin{equation}
\begin{aligned}
N_{k,m}(\alpha,\beta) = \bigcup_{q=0}^m \bigcup_{r=0}^m N_q(\alpha) \cap N_r(\beta)
\\
\text{ and } \\
{\mathfrak{I}}_m(\alpha,\beta)=\sum_{q=0}^m \sum_{r=0}^m  n^{qr}(\alpha,\beta)
\end{aligned}
\end{equation}

With these notations, let $s =|\Sigma|$, $n^{ij}(\alpha,\beta)$ can be computed using the following closed form. 
\begin{theorem}[~\cite{farhan2017efficient}] \label{thm_Intersection_ClosedForm}
	Given two $k$-mers $\alpha$ and $\beta$ such that $d(\alpha,\beta)=d$, we have that
	\begin{equation}
	    \begin{aligned}
	    n^{ij}(\alpha,\beta)= \sum_{t=0}^{\frac{i+j-d}{2}}{2d - i-j+2t\choose d-(i-t)} {d \choose i+j-2t-d} \times  
	    \\
	    (s -2)^{i+j-2t-d}  {k-d\choose t} (s-1)^t
	    \end{aligned}
	\end{equation}
\end{theorem} 

\begin{corollary}\label{thmIntersection}
	Runtime of computing ${\cal I}_d$ is $O(m^3)$, independent of $k$ and $|\Sigma|$.
\end{corollary}
This is so, because if $d(\alpha,\beta)=d$, ${\cal I}_d = \sum\limits_{q=0}^m \sum\limits_{r=0}^m n^{qr}(\alpha,\beta)$ and $n^{qr}(\alpha,\beta)$ can be computed in $O(m)$.

\subsection{Computing $M_i(X,Y)$}\label{subsec:computeMd}
Recall that given two sequences $X$ and $Y$, $M_i(X,Y)$ is the number of pairs of $k$-mers $(\alpha,\beta)$ such that $d(\alpha,\beta)=i$, where $\alpha\in X$ and $\beta\in Y$. Formally, the problem of computing $M_i(X,Y)$ is as follows: 
\begin{problem}\label{problem:1}
	Given $k$, $m$, and two sets of $k$-mers $S_X$ and $S_Y$ (set of $k$-mers extracted from the sequences $X$ and $Y$,  respectively) with $|S_X|=n_X$ and $|S_Y| = n_Y$. Let $t = \min\{2m,k\}$, for $0\leq i \leq t$ compute $$M_i(X,Y) = |\{(\alpha,\beta) \in S_X\times S_Y : d(\alpha,\beta) = i\}| $$
\end{problem}
Note that the brute force approach to compute $M_i(X,Y)$ requires $O(n_X\cdot n_Y\cdot k)$ comparisons. 
Let $\Q _k(j)$ denote the set of all $j$-sets of $\{1,\ldots,k\}$ (subsets of indices). For $\theta \in \Q _k(j)$ and a $k$-mer $\alpha$, let $\alpha|_{\theta}$ be the $j$-mer obtained by selecting the characters at the $j$ indices in $\theta$. Let $f_{\theta}(X,Y)$ be the number of pairs of $k$-mers in $S_X\times S_Y$ as follows; $$f_{\theta}(X,Y) = |\{(\alpha,\beta) \in S_X\times S_Y : d(\alpha|_{\theta},\beta|_{\theta}) = 0\}|.$$ We use the following important observations about $f_{\theta}$.
\begin{fact} For $0\leq i \leq k$ and $\theta \in \Q _k(k-i)$, if $d(\alpha|_{\theta},\beta|_{\theta}) = 0$, then $d(\alpha,\beta) \leq i$.
\end{fact}
\begin{fact}\label{fact6} For $0\leq i \leq k$ and $\theta \in \Q_k(k-i)$, $f_{\theta}(X,Y)$ can be computed in $O(kn\log n)$ time.
\end{fact}
\begin{proof}
This can be done by first lexicographically sorting the $k$-mers in each of $S_X$ and $S_Y$ by the indices in $\theta$. The pairs in $S_X\times S_Y$ that are the same at indices in $\theta$ can then be enumerated in one linear scan over the sorted lists. Let $n=n_X+n_Y$, the running time of this computation is $O(k(n+|\Sigma|))$ if we use counting sort (as in \cite{kuksa2009scalable}) or $O(kn\log n)$ for mergesort (since $\theta$ has $O(k)$ indices.) Since this procedure is repeated many times, we refer to this as the \textsc{sort-enumerate} subroutine.
\end{proof}
Define \beq\label{Eq:Fiformula} F_i(X,Y) = \sum_{\theta \in \Q_k(k-i)} f_{\theta}(X,Y).\eeq 

We can compute $M_i(X,Y)$ from $F_j(X,Y)$ using the following identity.
\begin{lemma} \beq \label{FalternateForm} F_i(X,Y) \;=\; \sum_{j=0}^i {k-j \choose k-i}M_j(X,Y).\eeq
\end{lemma}
\begin{proof}
Let $(\alpha,\beta)$ be a pair in $X\times Y$ that contributes to $M_j(X,Y)$, i.e. $d(\alpha,\beta) = j$. Then for every $\theta \in \Q_k(k-i)$ that has all indices within the $k-j$ positions where $\alpha$ and $\beta$ agree, the pair $(\alpha,\beta)$ is counted in $f_{\theta}(X,Y)$. The number of such $\theta$'s are ${k-j \choose k-i}$, hence $M_j(X,Y)$ is counted ${k-j \choose k-i}$ times in $F_i(X,Y)$, yielding the identity. 
	
\end{proof}
\begin{corollary}\label{corrMi}
	$M_i(X,Y)$ can readily be computed as: $$ M_i(X,Y) \;=\; F_i(X,Y) - \sum\limits_{j=0}^{i-1} {k-j \choose k-i}M_j(X,Y)$$
\end{corollary}

Next, we derive expressions for $M_i$ matrices for $1\leq i \leq \min\{2m,k\}$ (with values of $M_i(X,Y)$ for all pairs of sequences). In this alternate form it is easier to approximate the matrix $M_i$ and show that the resultant approximate kernel matrix indeed is positive semidefinite, as required for kernel based machine learning methods. 

For a sequence $X$ and $\theta \in \Q_k(j)$, let $\vec{u}_\theta(X)$ be a $|\Sigma|^j$ dimensional vector defined as: 
\begin{equation}
	\label{thetaRep} \vec{u}_\theta(X) = \left( \vec{u}_{\theta}(X)[\gamma]\right)_{\gamma \in \Sigma^j} 
	=     \left( \sum_{\alpha \in X} I(\alpha|_\theta,\gamma)\right)_{\gamma \in \Sigma^j} 
\end{equation}
where $I(\alpha|_\theta,\gamma)=1$, if $\alpha|_\theta = \gamma$.

It is easy to see that by definition $f_\theta(X,Y) = \langle \vec{u}_\theta(X),\vec{u}_\theta(X)\rangle$. Let $U_i(X)$ be the concatenation $\vec{u}_\theta(X)$ for all $\theta \in \Q_k(k-i)$. Again by definition of $F_i(X,Y)$ in \eqref{Eq:Fiformula} we have that $F_i(X,Y) = \langle U_i(X), U_i(Y)\rangle$.

Let $F_i$ and $M_i$ be $N\times N$ matrix with a row and column corresponding to each of the $N$ sequences, with values $F_i(X,Y)$ and $M_i(X,Y)$ for all pairs of sequences $X$ and $Y$. We get the matrix versions of Lemma \ref{FalternateForm} and Corollary \ref{corrMi}, i.e.  $$F_i \;=\; \sum_{j=0}^i {k-j \choose k-i}M_j \quad \text{ and } \quad M_i \;=\; F_i - \sum\limits_{j=0}^{i-1} {k-j \choose k-i}M_j$$

If $U_i$ is a matrix with $U_i(X)$'s as its $N$ columns, then by definition, $F_i = U_i^TU_i$, thus $F_i$ is a positive semidefinite matrix. Using Lemma \ref{FalternateForm} one easily verify for $0\leq i \leq \min\{2m,k\}$, the matrices $M_i$ are also positive semidefinite.

Note that for space and computational efficiency we use \ref{Eq:Fiformula} and Corollary \ref{corrMi} to compute $F_i$ and $M_i$ to compute  By definition, $F_i(X,Y)$ can be computed with ${k\choose k-i} ={k\choose i}$ $f_{\theta}$ computations. $K(X,Y|k,m)$ can be evaluated by \eqref{Eq:MK2} after computing $M_i(X,Y)$ (by \eqref{FalternateForm}) and ${\cal I}_i$ (by Corollary \ref{thmIntersection}) for $0\leq i\leq t$. The overall complexity of this strategy thus is $$\left(\sum_{i=0}^t {k\choose i} (k-i)(n\log n + n)\right) + O(n) = O(k\cdot 2^{k-1} \cdot (n\log n)).$$

Next, we give our sampling based approximate method for kernel computation. We select a random sample of index sets in $\Q_{k-i}$ for each $i$ and compute an estimate $\hat{F}_i^{xy}$ of $F_i(X,Y)$. These $\hat{F}_i$'s are used to compute estimates $\hat{M}_i$ of $M_i$. In matrix form, this corresponds to a random projection of the vectors $U_i(X)$ on the subspace spanned by the selected random index set. Thus, the resulting matrices $\hat{M}_i$'s are positive semidefinite, leading to positive semidefinite kernel matrix.

\begin{algorithm}[h!]
	\caption{: Approximate-Kernel($S_X$,$S_Y$,$k$,$m$,
	$\epsilon, \delta$) to estimate $K(X,Y|k,m)$}
	\label{algo:approxKernel}
	\begin{algorithmic}[1]
			\State ${\cal I}\gets \Call{zeros}{t+1}$ 
			\State $\hat{M} \gets \Call{zeros}{t+1}$
			\State $\hat{F} \gets \Call{zeros}{t+1}$
			\State Populate ${\cal I}$ using Corollary \ref{thmIntersection}
			\For{$i =0$ to $t$}
			\State $\mu_F \gets 0$
			\For{$\theta \in B_i$}
\State $\mu_F \gets \mu_F + \Call{sort-enumerate}{S_X,S_Y,k,\theta}$ \Comment{Application of Fact \ref{fact6}}
			\EndFor
			\State $\hat{F}[i] \gets \mu_F\cdot \frac{1}{|B_i|}{k\choose k-i}$
			\State $\hat{M}[i]\gets \hat{F}[i]$
			\For{$j = 0$ to $i-1$} \Comment{Application of Corollary \ref{corrMi}}
			\State $\hat{M}[i] \gets \hat{M}[i] - {k-j \choose k-i}\cdot \hat{M}[j]$
			\EndFor
			\EndFor
			\State $K' \gets \Call{sumproduct}{\hat{M},{\cal I}}$ \Comment{Applying Equation \eqref{Eq:MK2}}
			\State \Return $K'$
	\end{algorithmic}
\end{algorithm}

We randomly sample a collection $B_i$ of index-sets from $\c Q_{k-i}$, that Algorithm \ref{algo:approxKernel} uses to compute estimate $\hat{F}_i^{xy}$ for a pair of sequences $X,Y$. Using \eqref{Eq:Fiformula} to estimate $\hat{F}_i^{xy}$ for the randomly chosen $\theta\in B_i$. The estimated $\hat{F}_i^{xy}$'s are used to compute $\hat{M}_i^{xy}$ (estimates of $M_i(X,Y)$) using Corollary \ref{corrMi}. These $\hat{M}_i^{xy}\;$'s together with the pre-computed exact values of ${\cal I}_i$'s are used to compute our estimate, $K'(X,Y|k,m,\sigma,\epsilon,\delta)$ for the kernel value using \eqref{Eq:MK2}. The sample sizes (cardinalities of $B_i$'s) are chosen such that variance in the estimates are bounded above by $\sigma^2=\epsilon^2 \delta$, where $\epsilon$ and $\delta$ are user set parameters.

First, we give an analytical bound on the runtime of Algorithm \ref{algo:approxKernel} then we provide guarantees on its performance.
\begin{theorem}\label{runtime}
	Runtime of Algorithm \ref{algo:approxKernel} is $O(k^2 n\log n)$.
\end{theorem}
\begin{proof} 
	Let $B = \max_{i<t}\{|B_i|\}$. Observe that throughout the execution of the algorithm there are at most $tB$ computations of $f_{\theta}$, which by Fact \ref{fact6} needs $O(kn\log n)$ time. Since $B$ is an absolute constant and $t\leq k$, we get that the total runtime of the algorithm is $O(k^2n\log n)$.
\end{proof}
Let $K' = K'(X,Y|k,m,\epsilon,\delta)$ be our estimate (output of Algorithm \ref{algo:approxKernel}) for $K = K(X,Y|k,m)$.
\begin{theorem}\label{unbiasedEst}
	$K'$ is an unbiased estimator of the true kernel value, i.e. $E(K') = K$.
\end{theorem}
\begin{proof} For this, we need the following result, whose proof is deferred.
	\begin{lemma}\label{lem:unbiasedKernel} $E(\hat{M}_i^{xy}) = {M}_i(X,Y)$.
	\end{lemma}
	By Line 17 of Algorithm \ref{algo:approxKernel}, $E(K') =E( \sum_{i=0}^{t} {\cal I}_i \hat{M}_i^{xy}).$ Using the fact that ${\cal I}_i$'s are constants and Lemma \ref{lem:unbiasedKernel} we get that $$E(K') = \sum_{i=0}^{t} {\cal I}_i E(\hat{M}_i^{xy})= \sum_{i=0}^{\min\{2m,k\}} {\cal I}_i {M}_i(X,Y) = K.$$\end{proof} 
\begin{theorem}\label{error}
	For any $0< \epsilon, \delta < 1$, Algorithm \ref{algo:approxKernel} is an $(\epsilon {\cal I}_{max}, \delta)-$additive approximation algorithm, i.e. $Pr(|K-K'| \geq \epsilon {\cal I}_{max} ) < \delta$, where ${\cal I}_{max} = \max_{i}\{{\cal I}_i\}$.
\end{theorem}
Note that though ${\cal I}_{max}$ could be large, but it is only a fraction of one of the terms in summation for the kernel value $K(X,Y|k,m)$.
\begin{proof} 
	Let $\hat{F}_i^{xy}$ be our estimate for $F_i\l X,Y\r $. We use the following bound on the variance of $K'$ that is proved later.
	\begin{lemma}\label{lem:kernelVariance}
		$Var(K') \leq \delta(\epsilon\cdot{\cal I}_{max})^2.$
	\end{lemma}
By Lemma \ref{lem:unbiasedKernel} we have $E(K') = K$, hence by Lemma \ref{lem:kernelVariance}, $Pr[|K' - K|] \geq \epsilon {\cal I}_{max}$ is equivalent to $Pr[|K' - E(K')|] \geq \frac{1}{\sqrt{\delta}}\sqrt{Var(K')}$. By Chebychev's inequality, this latter probability is at most $\delta$. Therefore, Algorithm \ref{algo:approxKernel} is an $(\epsilon {\cal I}_{max}, \delta)-$additive approximation algorithm. \end{proof} 

\begin{proof}(Proof of Lemma \ref{lem:unbiasedKernel})
We prove it by induction on $i$. The base case ($i=0$) is  true as we compute $M'[0]$ exactly, i.e. $\hat{M}[0] = {M}_0(X,Y)$. Suppose $E(\hat{M}_j^{xy}) = {M}_i(X,Y)$ for $0\leq j \leq i-1$. After execution of Line 10 we get $$\hat{F}[i] = \frac{1}{|B_i|} \mu_F  {k\choose k-i} = \frac{1}{|B_i|}  \sum_{r=1}^{|B_i|} f_{\theta_r}(X,Y) {k\choose k-i},$$ where $\theta_r$ is the random $(k-i)$ index set. Since $\theta_r$ is chosen uniformly at random, we get that  
\beq\label{Eq:expectedOurF} 
E(\hat{F}[i]) = E(f_{\theta_r}(X,Y)) {k\choose k-i} = \dfrac{F_i(X,Y)}{{k\choose k-i}}  {k\choose k-i}
\eeq 

After the loop on Line 15 is executed we get that $E(\hat{M}[i]) = F_i(X,Y) - \sum\limits_{j=0}^{i-1} {k-j \choose k-i}E(\hat{M}_j^{xy})$. Using $E(\hat{M}_j^{xy})={M}_j(X,Y)$ (inductive hypothesis) in \eqref{FalternateForm} we get that $E(\hat{M}_i^{xy}) = {M}_i(X,Y)$. 
\end{proof}

\begin{proof} (Proof of Lemma \ref{lem:kernelVariance}) After execution of the inner loop in Algorithm \ref{algo:approxKernel}, we have $\hat{F}_i^{xy} =  \sum\limits_{j=0}^i {k-j \choose k-i}\hat{M}_j^{xy}.$
We use the following fact that follows from basic calculations. 
\begin{fact}\label{fact:varLinComb}
	Suppose $X_0,\ldots,X_t$ are random variables and let $S= \sum_{i=0}^t a_iX_i$, where $a_0,\ldots,a_t$ are constants. Then $$Var(S) = \sum_{i=0}^t a_i^2Var(X_i) + 2\sum_{i=0}^t\sum_{j=i+1}^t a_ia_jCov(X_i,X_j).$$
\end{fact} 
Using fact \ref{fact:varLinComb} and 
definitions of ${\cal I}_{max}$ and $\sigma$ we get that 

\begin{align}
    Var(K')&=\sum_{i=0}^t {{\cal I}_i}^2 Var( \hat{M}_i^{xy}) +2\sum_{i<j}^t{\cal I}_i{\cal I}_j Cov(\hat{M}_i^{xy}, \hat{M}_j^{xy}) \notag
    \\
    &\leq {\cal I}_{max}^2 \bigg[\sum_{i=0}^t Var(\hat{M}_i^{xy})  +2\sum_{i<j}^t Cov( \hat{M}_i^{xy}, \hat{M}_j^{xy})\bigg]  \notag
    \\ 
    &\leq {\cal I}_{max}^2 Var(\hat{F}_t^{xy}) \leq {\cal I}_{max}^2 \sigma^2 =\delta(\epsilon\cdot{\cal I}_{max})^2
    \end{align}

The last inequality follows from the following relation derived from the definition of $F_i'$ and Fact \ref{fact:varLinComb}.

\begin{equation}\label{Eq:VarourF}
    \begin{aligned}
    Var(\hat{F}_t^{xy})=\sum_{i=0}^t {k-i\choose k-t}^2 Var(\hat{M}_i^{xy}) 
    \\
    +2\sum_{i<j}^t{k-i\choose k-t}{k-j\choose k-t} Cov( \hat{M}_i^{xy}, \hat{M}_j^{xy})
    \end{aligned}
\end{equation}
\end{proof}

\begin{remark}
For reference, we call this kernel based method as \textit{Kernel Approximate} (or Kernel Approx.). We use $k=3$ and $m=0$ for this method, which is decided using standard validation set approach~\cite{validationSetApproach}.
\end{remark}

\section{Ordered Minimizer with Kernel (OMK)}~\label{sec_omk}
\begin{figure}[h!]
    \centering
    \includegraphics[scale = 0.2] {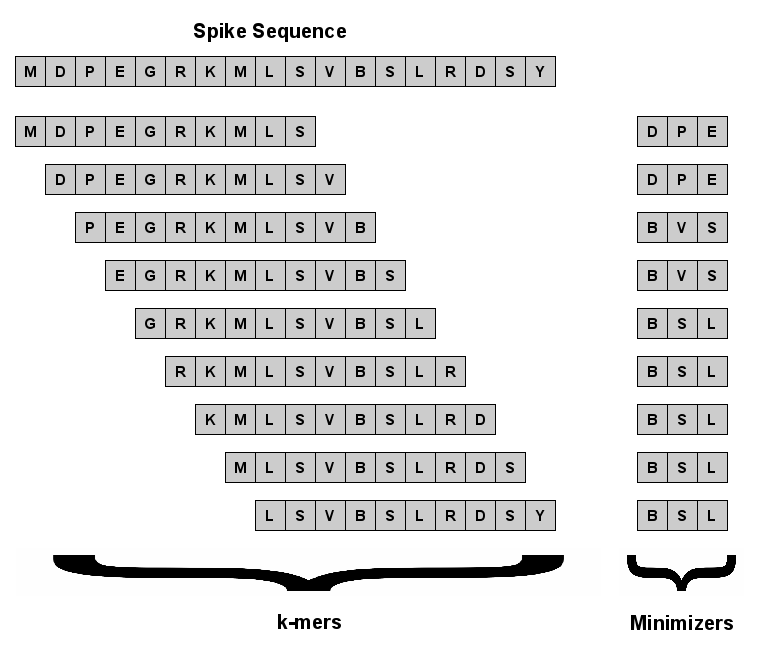}
    \caption{Example of $k$-mers and Minimizers in a spike sequence ``MDPEGRKMLSVBSLRDSY".}
    \label{fig_k_mers_minimizer}
\end{figure}

The original study of approximate kernel in~\cite{farhan2017efficient} (and the one that uses approximate kernel in our conference version of the paper~\cite{ali2021k}) uses the $k$-mers-based method to compute the approximate values for the kernel matrix. One problem with the $k$-mers is that there could be a lot of similar $k$-mers in a sequence that may not add any value to boost the predictive performance of the underlying machine learning (ML) algorithms. These redundant $k$-mers, however, could add computational overhead to the underlying processing. 
One solution to deal with this problem is to use Minimizers~\cite{roberts2004reducing} (also called $m$-mers), where $m<k$. 
The main idea of the minimizer ($m$-mers) is the following: Given a $k$-mer, an $m$-mer $\in$ $k$-mer is a sub-sequence, which is lexicographically smallest both in forward and reverse order of the $k$-mer (as given in Figure~\ref{fig_k_mers_minimizer}).
Since $m<k$, we ignore some of the amino acids (which were in $k$-mers but not in $m$-mers), which helps us to reduce computational overhead (as input size is reduced). The pseudocode to compute minimizers given a sequence is given in Algorithm~\ref{algo_minimizer}.
Although the notion of minimizer is previously used in the domain of
metagenomics~\cite{girotto2016metaprob}, it has not been used (to the best of our knowledge) for COVID-19 sequence classification.

\begin{algorithm}[h!]
	\caption{Minimizer Computation}
\label{algo_minimizer}
	\begin{algorithmic}[1]
	\State \textbf{Input:} Sequence $s$ and integer $k$ and $m$
	\State \textbf{Output:} Set of Minimizers
	
	\State minimizers $\gets \emptyset$
    \State queue $\gets []$ \Comment{$ \text{maintain queue of all m-mers }$}
    \State idx $\gets 0$  \Comment{ $ \text{index of the current minimizer}$}
     \For{$i = 1 \textup { to }|s|-k+1$}   
        \State kmer $\gets s[i:i+k]$ 

        \If{\textup{idx} $> 1$} 
            \State queue.dequeue 
            \State mmer $\gets$ $s[i+k-m:i+k]$  \Comment{$  \text{new m-mer}$}
            \State idx $\gets$ idx $- 1$  \Comment{$  \text{shift index of current minimizer}$}
            \State mmer $\gets$ min(mmer, reverse(mmer))  \Comment{$  \text{lexicographically smallest forw./rever.}$}
            \State queue.enqueue(mmer) 
            
            \If{\textup{mmer $<$ queue[idx]}}
                \State idx  $\gets k-m$ \Comment{$  \text{update minimizer with new m-mer}$}
             \EndIf
        \Else
                
                \State queue = [] \Comment{$  \text{reset the queue}$}
                \State idx = 0
                \For{$j = 1 \textup{ to } k-m+1$}
                    \State mmer $\gets$ kmer$[j:j+m]$ \Comment{$ \text{compute each m-mer}$}
                    \State mmer $\gets$ min(mmer, reverse(mmer))
                    \State queue.enqueue(mmer)
                    \If{\textup{mmer $<$ queue[idx]}}
                        \State idx $\gets$ $j$ \Comment{$ \text{index of current minimizer}$}
                    \EndIf
                \EndFor
        \EndIf
        
        \State minimizers $\leftarrow$ minimizers $\cup$ queue[idx] \Comment{$ \text{add current minimizer}$}
    \EndFor 
        \State return(minimizers)
	\end{algorithmic}
\end{algorithm}

To use the power of minimizer with the approximate kernel approach~\cite{farhan2017efficient}, we perform the following operations:
Given a sequence, we first compute a set of minimizers from the $k$-mers (where $m = 3$ and $k = 9$), see Figure~\ref{fig_k_mers_minimizer} for an example. We then concat those minimizers to make a new sequence (for reference, we call this new sequence as $s_{minimizer}$). That sequence is used as an input for the approximate kernel algorithm to compute the kernel matrix. Since the approximate kernel method operates by computing $k$-mers, the order of the amino acids in $s_{minimizer}$ is preserved by applying $k$-mers approach. This is the reason we call this method Ordered Minimizer with Kernel (OMK). After computing the kernel matrix, we use kernel PCA~\cite{hoffmann2007kernel} to compute the feature vector representation (we selected 50 principal components for our experiments) for the sequences and apply different machine learning tasks on the vectors, such as classification and clustering.

\section{Information Gain with Kernel (IGK)}\label{sec_igk}
One way to compute the importance of amino acids in a sequence is to use Information Gain (IG)~\cite{ali2021k}. The
IG of an amino acid position in terms of a class (variant) is defined as follows:
\begin{equation}
  \mbox{IG}(Class,~position) = H(Class) - H(Class ~|~ position)
  \label{eq:ig}
\end{equation}
where
\begin{equation}
  H(C) = \sum_{i \in C} -p_i \log p_i
\end{equation}
is the entropy of category $C$, and $p_i$ is the probability of element $i$ of category $C$.  Intuitively, the information gain of a given amino acid position tells us how much information this position provides in deciding the class (variant). Given a sequence, we first compute IG values for amino acids. We then select the amino acids with top IG values (top $243$ amino acids, which are selected using standard validation set approach) and use those amino acids only as input to the approximate kernel method. The approximate kernel approach computes the distance score for each pair of sequences based on the amino acids with top IG value only, and we then apply kernel PCA~\cite{hoffmann2007kernel} for non-kernel classifiers (we selected 50 principal components for our experiments using standard validation set approach) to compute the vector representation for the sequence and apply classification/clustering tasks. For reference, we call this method ``Information Gain with Kernel (IGK)".

\begin{remark}
Note that using less than $243$ amino acids gave worst classification results while having more that $243$ amino acids did not had any significant impact on the results.
\end{remark}

\section{Ordered Minimizer with Kernel and IG (OMK + IG)}\label{sec_omkig}
In this setting, we first compute the set of minimizers from the $k$-mers (where $m = 3$ and $k = 9$) for a given sequence as performed in Section~\ref{sec_omk} (see Figure~\ref{fig_k_mers_minimizer} for an example). After getting the minimizers, we concat them to make a single sequence ($s_{minimizer}$). The information gain logic is then applied on $s_{minimizer}$ to get the top amino acid (we selected the top $2184$ amino acids, in this case, using the standard validation set approach). Those top amino acids as given as input to the approximate kernel algorithm for the computation of kernel matrix. Then kernel PCA~\cite{hoffmann2007kernel} is applied for non-kernel classifiers (we selected 50 principal components for our experiments) to get the vector representation followed by classification/clustering methods.

\begin{remark}
Note that the idea of using IG in this paper is to reduce the dimensionality of sequence-based data, so that the kernel computation and hence classification/clustering tasks can be performed efficiently, while retaining good accuracies. Since IG give us the importance of each amino acid positions within sequences, we can take advantage of those \textit{importance scores} to extract the relevant amino acid positions and discard the rest. In this way, since only the important features (amino acids to be precise) are considered, we will get better embeddings, and hence better classification/clustering results (in less computational time) because the noise (if any) from the sequences is ignored.
\end{remark}

In summary, given the original spike sequence data, we have $1274$ amino acids in each sequences. For IGK, we select top $243$ amino acids. For OMK, given a spike sequence, we first compute $k$-mers (where $k=9$) and then from each $9$-mer, we compute $m$-mer (where $m-3$). Since number of $k$-mers in any sequence are $N - k + 1$ (where $N$ is the length of sequence), we will have $1274 - 9 + 1 = 1266$ $k$-mers. Now, since we compute an $m$-mer of length $3$ from each $k$-mer, we have $1266$ $m$-mers in total. If we concat those $m$-mers, we will get $1266 * 3 = 3798$ amino acids, which we called as ordered minimizer and use it as input to kernel method (we call this OMK). In the next step, we compute top $2184$ amino acids using IG from $3798$ amino acids and we call this method as OMK + IG.

\section{Experimental Setup}\label{sec_exp_setup}
We use $70$-$30 \%$ training and testing data split for experimentation. To tune the hyperparameters, we apply $5$ fold cross validation on the training data and then compute results on $30 \%$ unseen (held out) testing set. 
Experiments are conducted $5$ times with different random train-test data, and average $\pm$ standard deviation results are reported. 
All experiments are performed in python language on a Core i5 system with Windows 10 OS and 32 GB RAM.

\subsection{Dataset Statistics}\label{sec_dataset_stats}
We randomly sampled spike sequences from the largest known database of human SARS-CoV-2,  GISAID \footnote{\url{https://www.gisaid.org/}}. The sample size is $7000$ (each sequence is length $1274$). 
Since kernel-based algorithms require the kernel matrix saved in memory, in order to be able to perform experiments on a PC and avoid memory overflow, we use $7000$ sequences. Moreover, the computational overhead of the baseline methods at a larger scale hinders performing any meaningful comparison. Therefore, we chose $7000$ sequences.
Unlike the conference version, where only $5$ variants were considered in experiments, here we consider all $22$ variants. We used uniform random sampling, hence the proportions of variants in the sample are close to those in the whole dataset on the date of sampling. The proportion of variants on both datasets is given in Table~\ref{tbl_variant_information}. We repeated the experiments on two independent samples, referred to as GISAID-1 and GISAID-2.

\subsection{Data Visualization}
\begin{figure*}[h!]
\centering
\begin{subfigure}{.33\textwidth}
  \centering
  \includegraphics[scale=0.28]{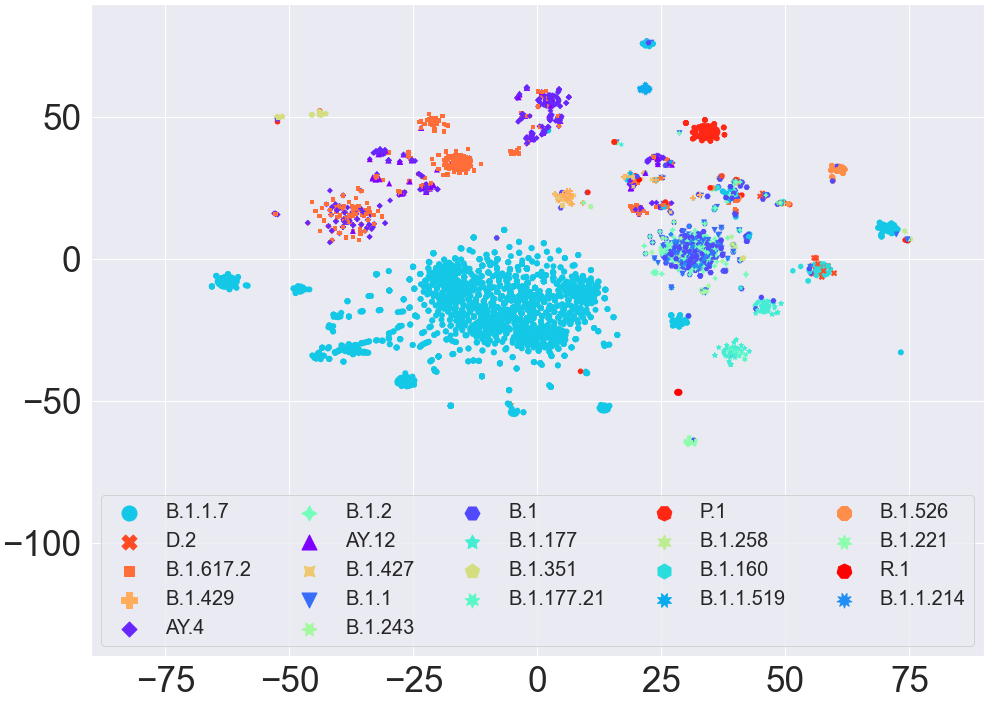}
  \caption{OHE}
  \label{}
\end{subfigure}%
\begin{subfigure}{.33\textwidth}
  \centering
  \includegraphics[scale=0.28]{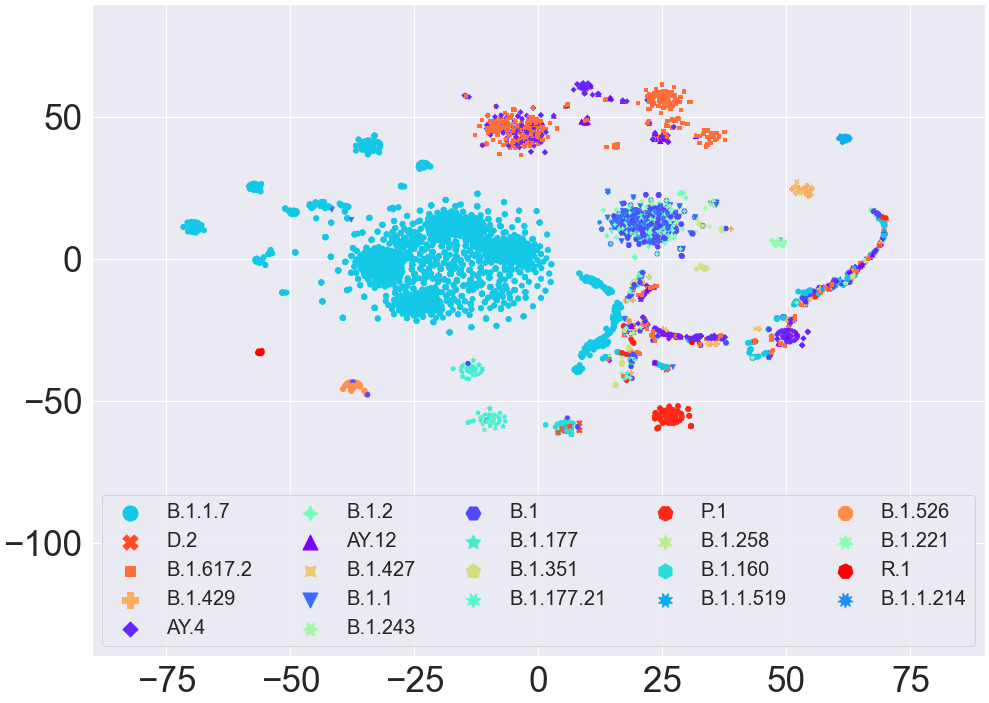}
  \caption{Spike2Vec}
  \label{}
\end{subfigure}%
\begin{subfigure}{.33\textwidth}
  \centering
  \includegraphics[scale=0.28]{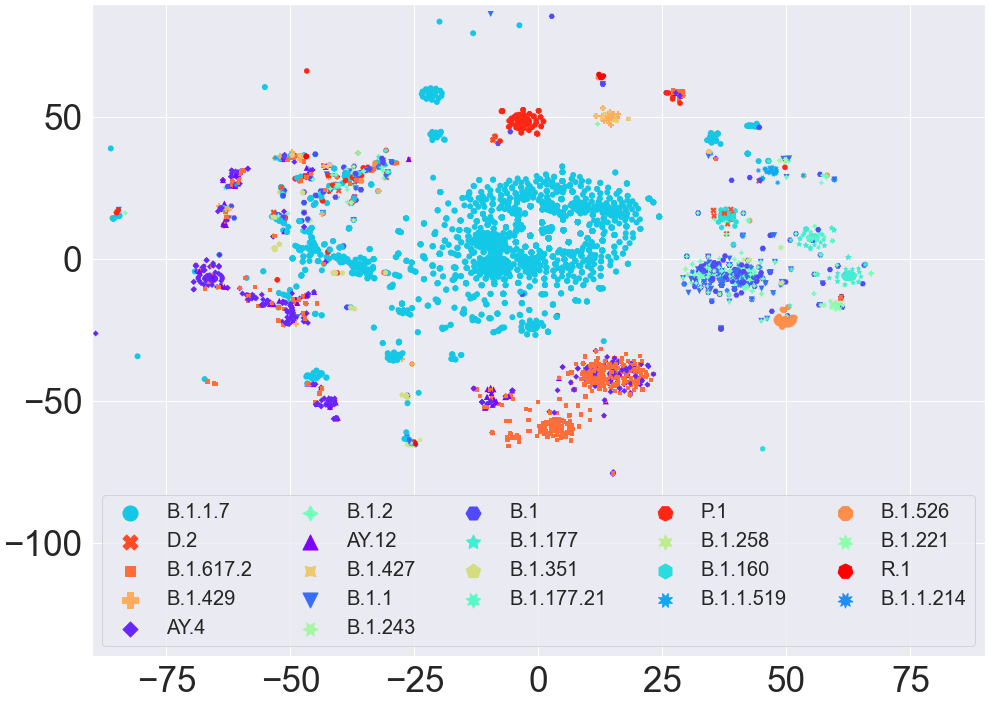}
  \caption{PWM2Vec}
  \label{}
\end{subfigure}%
\\
\begin{subfigure}{.33\textwidth}
  \centering
  \includegraphics[scale=0.28]{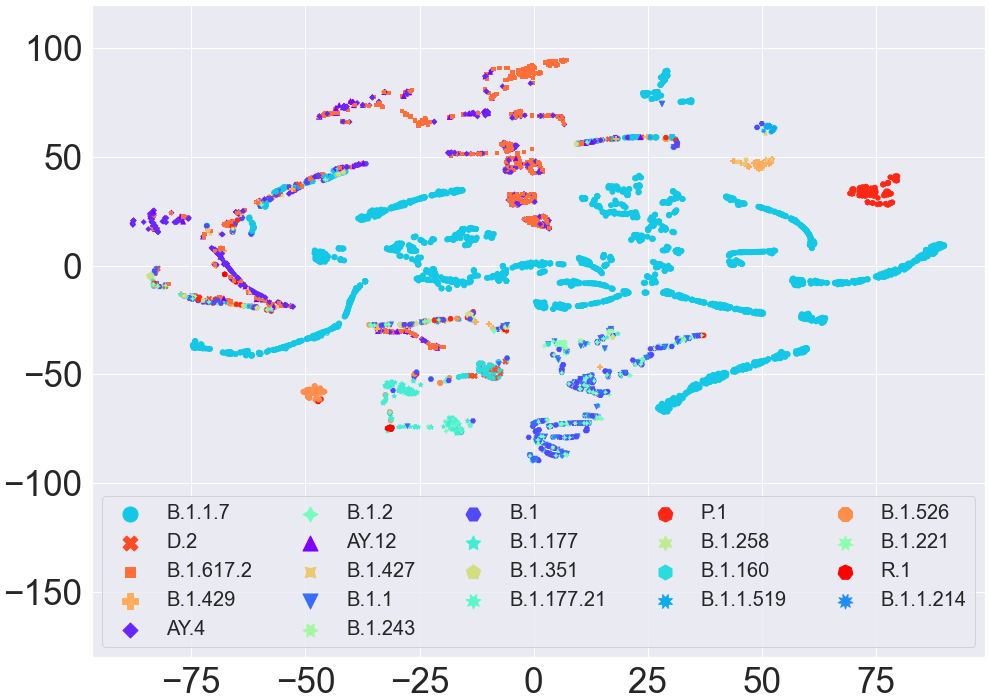}
  \caption{Kernel Approximate}
  \label{}
\end{subfigure}%
\begin{subfigure}{.33\textwidth}
  \centering
  \includegraphics[scale=0.28]{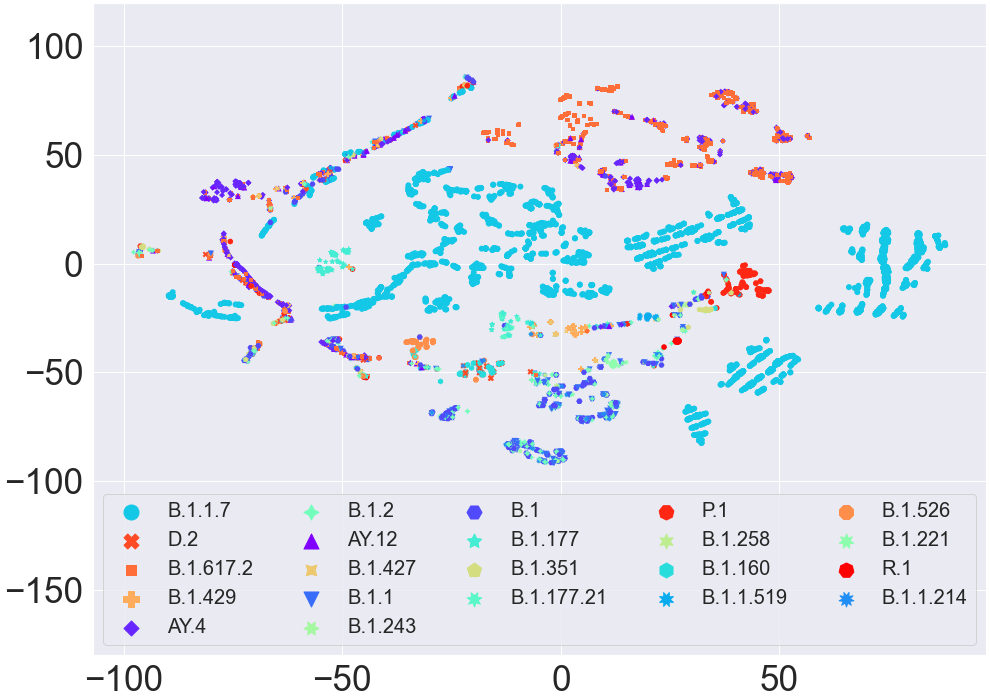}
  \caption{OMK}
  \label{}
\end{subfigure}%
\begin{subfigure}{.33\textwidth}
  \centering
  \includegraphics[scale=0.28]{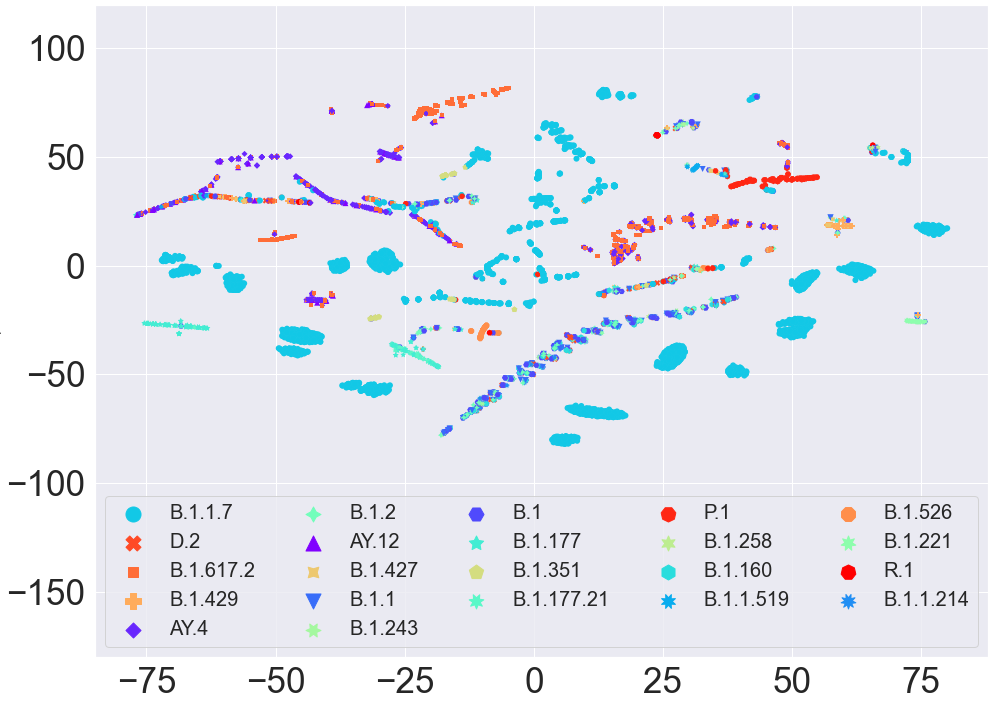}
  \caption{IGK}
  \label{}
\end{subfigure}%
\\
\begin{subfigure}{.33\textwidth}
  \centering
  \includegraphics[scale=0.28]{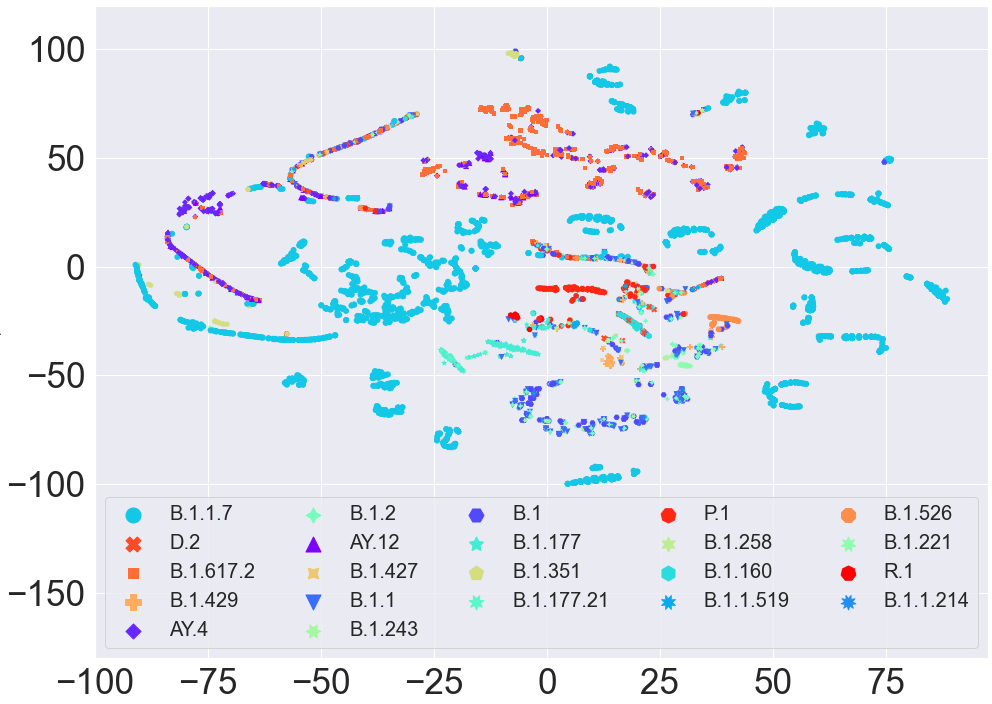}
  \caption{OMK + IG}
  \label{}
\end{subfigure}%
\caption{t-SNE plots for the SARS-CoV-2 dataset for different feature embeddings.}
\label{fig_all_tsne}
\end{figure*}

We use t-distributed stochastic neighbor embedding (t-SNE)
\cite{van2008visualizing} to evaluate the (hidden) patterns in the data. The t-SNE method maps input data to 2D real vectors that can be visualized using scatter plots.
The idea behind computing t-SNE plots is to see if the overall distribution of data is disturbed or remains the same when we use different embedding methods. The t-SNE plots for different embedding methods are shown in Figure~\ref{fig_all_tsne}. For the feature engineering-based methods (i.e. OHE, Spike2Vec, and PWM2Vec), we can see overlapping among different variants. However, for the kernel-based methods (for which embeddings were computed using kernel PCA), we can see a smaller (but comparatively pure) grouping of variants with less overlapping as compared to the feature engineering-based methods.

\begin{table}[h!]
  \centering
  \resizebox{0.49\textwidth}{!}{
  \begin{tabular}{@{\extracolsep{4pt}}p{0.9cm}lp{0.7cm}p{1.2cm} p{1.3cm} p{1.3cm}}
    \toprule
    \multirow{3}{*}{Lineages} & \multirow{3}{*}{Region} & \multirow{3}{*}{Labels} & \multirow{3}{1.2cm}{No. Mut. S/Gen.}  &  \multicolumn{2}{c}{No. of sequences}\\
    \cmidrule{5-6}
     &  &  &
    &  GISAID-1 &  GISAID-2 \\
    \midrule	\midrule	
    B.1.1.7 & UK~\cite{galloway2021emergence} &  Alpha & 8/17 & \hskip.1in 3369 & 3397 \\
    B.1.617.2  & India~\cite{yadav2021neutralization}  &  Delta &  8/17  & \hskip.1in 875 & 878 \\
    AY.4   & India~\cite{CDS_variantDef}  & Delta  &  - & \hskip.1in 593 & 516 \\
    B.1.2   & -  & -  & -  & \hskip.1in 333 & 350 \\
    B.1   &  & & & \hskip.1in 292  & 276 \\
    B.1.177   & Spain~\cite{hodcroft2020emergence} &  - & -  & \hskip.1in 243 & 281 \\
    P.1   & Brazil~\cite{naveca2021phylogenetic} &  Gamma &  10/21  & \hskip.1in 194  & 201 \\
    B.1.1   & -  & & -  & \hskip.1in 163  & 166 \\
    B.1.429   &  California  & Epsilon  & 3/5  & \hskip.1in 107  & 142 \\
    B.1.526   & New York~\cite{west2021detection}  &  Iota & 6/16 & \hskip.1in 104 & 82 \\
    AY.12   & India~\cite{CDS_variantDef}  & Delta  & -  & \hskip.1in 101 & 82 \\
    B.1.160   & -  & -  &  - & \hskip.1in 92 & 88 \\
    B.1.351  & South Africa~\cite{galloway2021emergence}  &  Beta & 9/21& \hskip.1in 81 & 62 \\
    B.1.427   & California~\cite{zhang2021emergence}  & Epsilon  &  3/5 & \hskip.1in 65 & 62 \\
    B.1.1.214   & -  & -  &  - & \hskip.1in 64 & 64 \\
    B.1.1.519   & -  & -  & -  & \hskip.1in 56 & 88 \\
    D.2   &  - &  - &  - & \hskip.1in 55 & 45 \\
    B.1.221   & -  & -  & -  & \hskip.1in 52 & 41 \\
    B.1.177.21   & -  &  - & -  & \hskip.1in 47 & 56 \\
    B.1.258   & -  & -  & -  & \hskip.1in 46 & 42 \\
    B.1.243   & -  & -  & -  & \hskip.1in 36 & 40 \\
    R.1   & -  & -  & -  & \hskip.1in 32 & 41 \\
    \midrule
    Total & -  & -  & -  & \hskip.1in 7000 & 7000 \\
    \bottomrule
  \end{tabular}
  }
  \caption{Dataset statistics for $22$ variants. The character `-' means that information not available.}
  \label{tbl_variant_information}
\end{table}

\subsection{Baseline Models}
In this section, we introduce the baseline (one-hot encoding) and the state-of-the-art (SOTA) approaches (Spike2Vec and PWM2Vec) that we used for comparison with our models.

\subsubsection{One-Hot Embedding (OHE)~\cite{kuzmin2020machine}}
A fixed-length numerical feature vector, OHE~\cite{ali2021k,kuzmin2020machine} generates 0-1 vector on the character's position in the sequence given $\Sigma$. The 0-1 vectors for all characters are concatenated to make a single vector for a given sequence. The length of the feature vector, in this case, is $30576$.

\subsubsection{Spike2Vec~\cite{ali2021spike2vec}}
Spike2Vec is recently proposed in ~\cite{ali2021spike2vec} for spike sequence classification. 
Given a sequence, Spike2Vec computes $N$ $k$-mers, where $N= L - k + 1$ ($L$ is the length of the spike sequence and $k=3$ as given in~\cite{ali2021spike2vec}). 
After generating the k-mers for a spike sequence, the count of each k-mer is used to get the frequency vector. The length of Spike2Vec-based embedding for each spike sequence is $\vert \Sigma \vert^k = 13824$.

\subsubsection{PWM2Vec~\cite{ali2022pwm2vec}}
A method combining the power of k-mers and position weight matrix~\cite{stormo-1982-pwm}, PWM2Vec is proposed in~\cite{ali2022pwm2vec}.
PWM2Vec assigns different weight to each k-mer (where $k=9$) in the feature vector depending on the values of the characters in the position weight matrix. 
The length of PWM2Vec-based embedding for each spike sequence is $1265$.

\subsection{Evaluation Metrics for Classification}
Various ML algorithms have been utilized for the
classification task. K-PCA output, which is $50$ components fed to
different classifiers for prediction purposes. We use Support Vector
Machine (SVM), Naive Bayes (NB), Multi-Layer Perceptron (MLP),
K-Nearest Neighbour (KNN) (with $K = 5$), Random Forest (RF), Logistic
Regression (LR), and Decision Tree (DT) classifiers.  
The evaluation metrics that we are using are average accuracy, precision, recall, weighted and macro F1, and ROC area under the curve (AUC).

\subsection{Evaluation Metrics for Clustering}
To perform the clustering on the data, we use the simple $k$-means algorithm.
To evaluate the performance of $k$-means, we use
the following internal evaluation metrics:

\textbf{Silhouette Coefficient}~\cite{rousseeuw1987silhouettes}: Given
a vector $v$, it measures the similarity of $v$ to its own cluster
(cohesion) compared to the other clusters (separation). Its value
range from $-1$ to $1$ where upper bound $1$ indicates best possible clustering and lower bound $-1$ shows worst possible clustering.

\textbf{Calinski-Harabasz Score}~\cite{calinski1974dendrite}: is the
ratio between the inter-cluster dispersion and the between-cluster
dispersion. The higher the value of this score, the higher the clustering
performance.

\textbf{Davies-Bouldin Score}~\cite{davies1979cluster}: It validates
the clustering schemes by measuring the similarity between
clusters. The ratio of distances within-cluster to between clusters is
referred to as similarity. Unlike the previous metrics, a lower
Davies-Bouldin Score indicates a better clustering performance, and
its lower bound is $0$.

\subsubsection{Elbow Method for k-means}
\begin{figure}
    \centering
    \includegraphics[scale = 0.4]{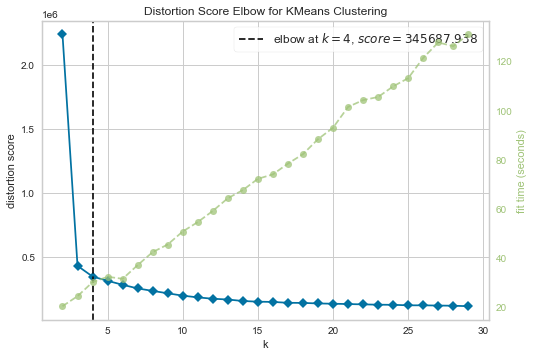}
    \caption{Elbow method to find the optimal number of clusters.}
    \label{fig_elbow_method}
\end{figure}

To get the optimal number of clusters, we use the Elbow method~\cite{satopaa2011finding,ali2021effective}. The main idea of the elbow method is to compute clusterings and evaluate the trade-off between two metrics, namely runtime and the sum of squared error (distortion score). The clustering having the optimal value for both metrics is selected as
the ideal number of clusters. The optimal number of clusters in this case is $4$ (see Figure~\ref{fig_elbow_method}).

\section{Results and Discussion}\label{sec_results}
In this section, we report classification and clustering results using our proposed model, baseline, and SOTA methods.

\subsection{Classification Results}
The classification results (average $\pm$ standard deviation values of $5$ runs) for GISAID-1 dataset are given in Table~\ref{tbl_classification_gisaid1}. We can observe that OMK+IG outperforms all other methods, including the baseline and SOTA, in terms of all evaluation metrics. Similar behavior is observed for GISAID-2 dataset (see Table~\ref{tbl_classification_gisaid2}). From these observations, we can conclude that using the domain knowledge from IG along with the power of minimizers, we can improve the classification performance of the approximate kernel.

\begin{table*}[h!]
    \centering
    \resizebox{0.99\textwidth}{!}{
    \begin{tabular}{p{1.2cm}cp{1.9cm}p{1.7cm}p{1.9cm}p{1.7cm}p{1.7cm}p{1.7cm}|p{2.1cm}}
    \toprule
        & & Acc. & Prec. & Recall & F1 (Weig.) & F1 (Macro) & ROC AUC & Train Time (Sec.) \\
        \midrule \midrule
        
        \multirow{7}{1.2cm}{OHE}
          &  SVM  & 0.83 $\pm$ 0.0019 & 0.83 $\pm$ 0.0053 & 0.83 $\pm$ 0.0019 & 0.82 $\pm$ 0.0030 & 0.67 $\pm$ 0.0140 & 0.82 $\pm$ 0.0047 & 301.53 $\pm$ 0.2618 \\
          &  NB  & 0.64 $\pm$ 0.0085 & 0.75 $\pm$ 0.0084 & 0.64 $\pm$ 0.0096 & 0.65 $\pm$ 0.0089 & 0.48 $\pm$ 0.0155 & 0.75 $\pm$ 0.0102 & 18.9 $\pm$ 0.2816 \\
          &  MLP  & 0.79 $\pm$ 0.0045 & 0.81 $\pm$ 0.0604 & 0.79 $\pm$ 0.0045 & 0.78 $\pm$ 0.0060 & 0.61 $\pm$ 0.0202 & 0.79 $\pm$ 0.0079 & 164.05 $\pm$ 0.0164 \\
          &  KNN  & 0.8 $\pm$ 0.0116 & 0.81 $\pm$ 0.0074 & 0.8 $\pm$ 0.0116 & 0.79 $\pm$ 0.0099 & 0.6 $\pm$ 0.0287 & 0.79 $\pm$ 0.0158 & 498.46 $\pm$ 1.4808 \\
          &  RF  & 0.82 $\pm$ 0.0066 & 0.82 $\pm$ 0.0096 & 0.82 $\pm$ 0.0066 & 0.8 $\pm$ 0.0077 & 0.64 $\pm$ 0.0142 & 0.8 $\pm$ 0.0053 & 29.52 $\pm$ 0.0147 \\
          &  LR  & 0.83 $\pm$ 0.0048 & 0.83 $\pm$ 0.0050 & 0.83 $\pm$ 0.0048 & 0.82 $\pm$ 0.0065 & 0.67 $\pm$ 0.0344 & 0.81 $\pm$ 0.0173 & 70.07 $\pm$ 0.0442 \\
          &  DT  & 0.83 $\pm$ 0.0100 & 0.83 $\pm$ 0.0112 & 0.83 $\pm$ 0.0100 & 0.82 $\pm$ 0.0103 & 0.68 $\pm$ 0.0242 & 0.82 $\pm$ 0.0144 & 6.25 $\pm$ 0.0120 \\

        \midrule
        \multirow{7}{1.2cm}{Spike2Vec}
          &  SVM  & 0.85 $\pm$ 0.0017 & 0.84 $\pm$ 0.0047 & 0.85 $\pm$ 0.0017 & 0.83 $\pm$ 0.0027 & 0.68 $\pm$ 0.0126 & 0.83 $\pm$ 0.0043 & 230.57 $\pm$ 0.2356 \\
          &  NB  & 0.35 $\pm$ 0.0077 & 0.73 $\pm$ 0.0076 & 0.35 $\pm$ 0.0087 & 0.45 $\pm$ 0.0080 & 0.45 $\pm$ 0.0140 & 0.72 $\pm$ 0.0092 & 12.54 $\pm$ 0.2534 \\
          &  MLP  & 0.79 $\pm$ 0.0040 & 0.81 $\pm$ 0.0544 & 0.79 $\pm$ 0.0040 & 0.8 $\pm$ 0.0054 & 0.58 $\pm$ 0.0182 & 0.79 $\pm$ 0.0071 & 65.79 $\pm$ 0.0147 \\
          &  KNN  & 0.82 $\pm$ 0.0104 & 0.82 $\pm$ 0.0067 & 0.82 $\pm$ 0.0104 & 0.81 $\pm$ 0.0089 & 0.6 $\pm$ 0.0258 & 0.78 $\pm$ 0.0142 & 115.85 $\pm$ 1.3327 \\
          &  RF  & 0.85 $\pm$ 0.0059 & 0.84 $\pm$ 0.0086 & 0.85 $\pm$ 0.0059 & 0.83 $\pm$ 0.0069 & 0.66 $\pm$ 0.0128 & 0.82 $\pm$ 0.0047 & 15.62 $\pm$ 0.0133 \\
          &  LR  & 0.85 $\pm$ 0.0044 & 0.85 $\pm$ 0.0045 & 0.85 $\pm$ 0.0044 & 0.84 $\pm$ 0.0058 & 0.68 $\pm$ 0.0310 & 0.83 $\pm$ 0.0156 & 50.74 $\pm$ 0.0398 \\
          &  DT  & 0.85 $\pm$ 0.0090 & 0.85 $\pm$ 0.0101 & 0.85 $\pm$ 0.0090 & 0.84 $\pm$ 0.0092 & 0.67 $\pm$ 0.0218 & 0.82 $\pm$ 0.0130 & 3.19 $\pm$ 0.0108 \\

         \midrule
        \multirow{7}{1.2cm}{PWM2Vec}
          &  SVM  & 0.82 $\pm$ 0.0015 & 0.83 $\pm$ 0.0042 & 0.82 $\pm$ 0.0015 & 0.81 $\pm$ 0.0024 & 0.63 $\pm$ 0.0112 & 0.81 $\pm$ 0.0038 & 173.89 $\pm$ 0.2095 \\
          &  NB  & 0.51 $\pm$ 0.0068 & 0.61 $\pm$ 0.0068 & 0.51 $\pm$ 0.0077 & 0.53 $\pm$ 0.0071 & 0.17 $\pm$ 0.0124 & 0.62 $\pm$ 0.0082 & 1.17 $\pm$ 0.2253 \\
          &  MLP  & 0.8 $\pm$ 0.0036 & 0.78 $\pm$ 0.0483 & 0.8 $\pm$ 0.0036 & 0.78 $\pm$ 0.0048 & 0.53 $\pm$ 0.0162 & 0.77 $\pm$ 0.0063 & 24.4 $\pm$ 0.0131 \\
          &  KNN  & 0.77 $\pm$ 0.0093 & 0.79 $\pm$ 0.0059 & 0.77 $\pm$ 0.0093 & 0.76 $\pm$ 0.0079 & 0.55 $\pm$ 0.0230 & 0.76 $\pm$ 0.0127 & 10.55 $\pm$ 1.1846 \\
          &  RF  & 0.83 $\pm$ 0.0053 & 0.83 $\pm$ 0.0077 & 0.83 $\pm$ 0.0053 & 0.82 $\pm$ 0.0061 & 0.63 $\pm$ 0.0113 & 0.8 $\pm$ 0.0042 & 13.54 $\pm$ 0.0118 \\
          &  LR  & 0.82 $\pm$ 0.0039 & 0.81 $\pm$ 0.0040 & 0.82 $\pm$ 0.0039 & 0.81 $\pm$ 0.0052 & 0.62 $\pm$ 0.0276 & 0.8 $\pm$ 0.0139 & 40.81 $\pm$ 0.0353 \\
          &  DT  & 0.8 $\pm$ 0.0080 & 0.81 $\pm$ 0.0090 & 0.8 $\pm$ 0.0080 & 0.8 $\pm$ 0.0082 & 0.59 $\pm$ 0.0193 & 0.79 $\pm$ 0.0116 & 2.63 $\pm$ 0.0096 \\

         \midrule
        \multirow{7}{1.2cm}{Kernel Approx.}
          &  SVM   & 0.84 $\pm$ 0.0016 & 0.83 $\pm$ 0.0045 & 0.84 $\pm$ 0.0016 & 0.82 $\pm$ 0.0026 & 0.63 $\pm$ 0.0120 & 0.81 $\pm$ 0.0040 & 7.35 $\pm$ 0.2239 \\
          &  NB    & 0.75 $\pm$ 0.0073 & 0.82 $\pm$ 0.0072 & 0.75 $\pm$ 0.0082 & 0.77 $\pm$ 0.0076 & 0.6 $\pm$ 0.0133 & 0.82 $\pm$ 0.0088 & 0.17 $\pm$ 0.2408 \\
          &  MLP   & 0.83 $\pm$ 0.0038 & 0.82 $\pm$ 0.0517 & 0.83 $\pm$ 0.0038 & 0.82 $\pm$ 0.0052 & 0.62 $\pm$ 0.0173 & 0.81 $\pm$ 0.0068 & 12.65 $\pm$ 0.0140 \\
          &  KNN   & 0.82 $\pm$ 0.0099 & 0.82 $\pm$ 0.0063 & 0.82 $\pm$ 0.0099 & 0.82 $\pm$ 0.0084 & 0.62 $\pm$ 0.0245 & 0.79 $\pm$ 0.0135 & 0.32 $\pm$ 1.2661 \\
          &  RF    & 0.84 $\pm$ 0.0056 & 0.84 $\pm$ 0.0082 & 0.84 $\pm$ 0.0056 & 0.83 $\pm$ 0.0066 & 0.66 $\pm$ 0.0121 & 0.82 $\pm$ 0.0045 & 1.46 $\pm$ 0.0126 \\
          &  LR    & 0.84 $\pm$ 0.0041 & 0.84 $\pm$ 0.0042 & 0.84 $\pm$ 0.0041 & 0.82 $\pm$ 0.0055 & 0.62 $\pm$ 0.0294 & 0.81 $\pm$ 0.0148 & 1.86 $\pm$ 0.0378 \\
          &  DT    & 0.82 $\pm$ 0.0086 & 0.82 $\pm$ 0.0096 & 0.82 $\pm$ 0.0086 & 0.82 $\pm$ 0.0088 & 0.63 $\pm$ 0.0207 & 0.82 $\pm$ 0.0124 & 0.24 $\pm$ 0.0102 \\
           \midrule
        \multirow{7}{1.2cm}{OMK}
          &  SVM  & 0.85 $\pm$ 0.0015 & 0.83 $\pm$ 0.0041 & 0.85 $\pm$ 0.0015 & 0.83 $\pm$ 0.0023 & 0.62 $\pm$ 0.0110 & 0.81 $\pm$ 0.0037 & 33.9 $\pm$ 0.2053 \\
          &  NB   & 0.74 $\pm$ 0.0067 & 0.8 $\pm$ 0.0066 & 0.74 $\pm$ 0.0075 & 0.76 $\pm$ 0.0070 & 0.59 $\pm$ 0.0122 & 0.8 $\pm$ 0.0080 & 0.13 $\pm$ 0.2208 \\
          &  MLP  & 0.83 $\pm$ 0.0035 & 0.82 $\pm$ 0.0474 & 0.83 $\pm$ 0.0035 & 0.82 $\pm$ 0.0047 & 0.61 $\pm$ 0.0158 & 0.8 $\pm$ 0.0062 & 21.77 $\pm$ 0.0128 \\
          &  KNN  & 0.81 $\pm$ 0.0091 & 0.81 $\pm$ 0.0058 & 0.81 $\pm$ 0.0091 & 0.8 $\pm$ 0.0077 & 0.63 $\pm$ 0.0225 & 0.8 $\pm$ 0.0124 & 0.31 $\pm$ 1.1609 \\
          &  RF   & 0.862 $\pm$ 0.0052 & 0.85 $\pm$ 0.0075 & 0.862 $\pm$ 0.0052 & 0.84 $\pm$ 0.0060 & 0.67 $\pm$ 0.0111 & 0.83 $\pm$ 0.0041 & 1.54 $\pm$ 0.0116 \\
          &  LR   & 0.85 $\pm$ 0.0038 & 0.84 $\pm$ 0.0039 & 0.85 $\pm$ 0.0038 & 0.83 $\pm$ 0.0051 & 0.63 $\pm$ 0.0270 & 0.81 $\pm$ 0.0136 & 2.99 $\pm$ 0.0346 \\
          &  DT   & 0.83 $\pm$ 0.0078 & 0.83 $\pm$ 0.0088 & 0.83 $\pm$ 0.0078 & 0.82 $\pm$ 0.0080 & 0.63 $\pm$ 0.0190 & 0.81 $\pm$ 0.0113 & 0.23 $\pm$ 0.0094 \\

          \midrule
        \multirow{7}{1.2cm}{IGK}
           &  SVM  & 0.85 $\pm$ 0.0018 & 0.84 $\pm$ 0.0051 & 0.85 $\pm$ 0.0018 & 0.83 $\pm$ 0.0029 & 0.6 $\pm$ 0.0136 & 0.8 $\pm$ 0.0046 & 3.23 $\pm$ 0.2540 \\
          &  NB  & 0.74 $\pm$ 0.0083 & 0.82 $\pm$ 0.0082 & 0.74 $\pm$ 0.0093 & 0.76 $\pm$ 0.0087 & 0.58 $\pm$ 0.0151 & 0.8 $\pm$ 0.0099 & 0.1 $\pm$ 0.2731 \\
          &  MLP  & 0.83 $\pm$ 0.0043 & 0.82 $\pm$ 0.0586 & 0.83 $\pm$ 0.0043 & 0.81 $\pm$ 0.0059 & 0.59 $\pm$ 0.0196 & 0.79 $\pm$ 0.0077 & 9.96 $\pm$ 0.0159 \\
          &  KNN  & 0.82 $\pm$ 0.0113 & 0.82 $\pm$ 0.0072 & 0.82 $\pm$ 0.0113 & 0.81 $\pm$ 0.0096 & 0.59 $\pm$ 0.0278 & 0.79 $\pm$ 0.0153 & 0.34 $\pm$ 1.4364 \\
          &  RF  & 0.84 $\pm$ 0.0064 & 0.83 $\pm$ 0.0093 & 0.84 $\pm$ 0.0064 & 0.82 $\pm$ 0.0074 & 0.59 $\pm$ 0.0138 & 0.8 $\pm$ 0.0051 & 1.36 $\pm$ 0.0143 \\
          &  LR  & 0.85 $\pm$ 0.0047 & 0.84 $\pm$ 0.0048 & 0.85 $\pm$ 0.0047 & 0.83 $\pm$ 0.0063 & 0.61 $\pm$ 0.0334 & 0.8 $\pm$ 0.0168 & 1.7 $\pm$ 0.0428 \\
          &  DT  & 0.83 $\pm$ 0.0097 & 0.82 $\pm$ 0.0109 & 0.83 $\pm$ 0.0097 & 0.81 $\pm$ 0.0100 & 0.58 $\pm$ 0.0234 & 0.79 $\pm$ 0.0140 & 0.21 $\pm$ 0.0116 \\
         \midrule
         \multirow{7}{1.2cm}{OMK + IG}
           &  SVM  &  \textbf{0.867}  $\pm$ 0.0016 & 0.85 $\pm$ 0.0045 &  \textbf{0.868}  $\pm$ 0.0016 &  \textbf{0.85}  $\pm$ 0.0025 & 0.66 $\pm$ 0.0119 & 0.83 $\pm$ 0.0040 & 20.83 $\pm$ 0.2216 \\
          &  NB  & 0.75 $\pm$ 0.0072 & 0.83 $\pm$ 0.0072 & 0.75 $\pm$ 0.0082 & 0.77 $\pm$ 0.0076 & 0.61 $\pm$ 0.0131 & 0.82 $\pm$ 0.0087 &  \textbf{0.09} $\pm$ 0.2384 \\
          &  MLP  & 0.84 $\pm$ 0.0038 & 0.84 $\pm$ 0.0511 & 0.84 $\pm$ 0.0038 & 0.83 $\pm$ 0.0051 & 0.65 $\pm$ 0.0171 & 0.83 $\pm$ 0.0067 & 13.26 $\pm$ 0.0138 \\
          &  KNN  & 0.83 $\pm$ 0.0098 & 0.84 $\pm$ 0.0063 & 0.83 $\pm$ 0.0098 & 0.83 $\pm$ 0.0084 & 0.65 $\pm$ 0.0243 & 0.81 $\pm$ 0.0134 & 0.31 $\pm$ 1.2534 \\
          &  RF  & 0.864 $\pm$ 0.0056 &  \textbf{0.86}  $\pm$ 0.0081 & 0.865 $\pm$ 0.0056 & 0.84 $\pm$ 0.0065 &  \textbf{0.69}  $\pm$ 0.0120 &  \textbf{0.84}  $\pm$ 0.0045 & 1.26 $\pm$ 0.0125 \\
          &  LR  & 0.865 $\pm$ 0.0041 & 0.85 $\pm$ 0.0042 & 0.86 $\pm$ 0.0041 & 0.84 $\pm$ 0.0055 & 0.63 $\pm$ 0.0292 & 0.82 $\pm$ 0.0147 & 2.08 $\pm$ 0.0374 \\
          &  DT  & 0.84 $\pm$ 0.0085 & 0.84 $\pm$ 0.0095 & 0.84 $\pm$ 0.0085 & 0.84 $\pm$ 0.0087 & 0.65 $\pm$ 0.0205 & 0.83 $\pm$ 0.0122 & 0.19 $\pm$ 0.0101 \\

        \bottomrule
    \end{tabular}
    }
    \caption{Average $\pm$ standard deviation classification results for  GISAID-1 dataset. Best average values are shown in bold.}
    \label{tbl_classification_gisaid1}
\end{table*}

\begin{table*}[h!]
    \centering
    \resizebox{0.99\textwidth}{!}{
    \begin{tabular}{p{1.2cm}cp{1.9cm}p{1.7cm}p{1.9cm}p{1.7cm}p{1.7cm}p{1.7cm}|p{2.1cm}}
    \toprule
        & & Acc. & Prec. & Recall & F1 (Weig.) & F1 (Macro) & ROC AUC & Train Time (Sec.) \\
        \midrule \midrule
        
        \multirow{7}{1.2cm}{OHE}
          &  SVM  & 0.84 $\pm$ 0.0017 & 0.84 $\pm$ 0.0046 & 0.84 $\pm$ 0.0018 & 0.83 $\pm$ 0.0028 & 0.6 $\pm$ 0.0138 & 0.8 $\pm$ 0.0045 & 285.83 $\pm$ 0.2511 \\
          &  NB  & 0.65 $\pm$ 0.0077 & 0.79 $\pm$ 0.0074 & 0.65 $\pm$ 0.0093 & 0.66 $\pm$ 0.0083 & 0.47 $\pm$ 0.0153 & 0.76 $\pm$ 0.0098 & 18.89 $\pm$ 0.2700 \\
          &  MLP  & 0.83 $\pm$ 0.0040 & 0.81 $\pm$ 0.0526 & 0.83 $\pm$ 0.0043 & 0.81 $\pm$ 0.0056 & 0.57 $\pm$ 0.0199 & 0.79 $\pm$ 0.0076 & 107.92 $\pm$ 0.1500 \\
          &  KNN  & 0.82 $\pm$ 0.0104 & 0.82 $\pm$ 0.0064 & 0.82 $\pm$ 0.0113 & 0.81 $\pm$ 0.0092 & 0.62 $\pm$ 0.0283 & 0.79 $\pm$ 0.0151 & 501.72 $\pm$ 2.3541 \\
          &  RF  & 0.85 $\pm$ 0.0059 & 0.85 $\pm$ 0.0083 & 0.85 $\pm$ 0.0064 & 0.83 $\pm$ 0.0071 & 0.63 $\pm$ 0.0140 & 0.81 $\pm$ 0.0050 & 29.84 $\pm$ 0.0985 \\
          &  LR  & 0.85 $\pm$ 0.0044 & 0.82 $\pm$ 0.0043 & 0.85 $\pm$ 0.0047 & 0.82 $\pm$ 0.0060 & 0.57 $\pm$ 0.0340 & 0.79 $\pm$ 0.0166 & 65.87 $\pm$ 0.1074 \\
          &  DT  & 0.83 $\pm$ 0.0090 & 0.82 $\pm$ 0.0098 & 0.83 $\pm$ 0.0097 & 0.82 $\pm$ 0.0095 & 0.6 $\pm$ 0.0239 & 0.8 $\pm$ 0.0138 & 6.49 $\pm$ 0.0824 \\

        \midrule
        \multirow{7}{1.2cm}{Spike2Vec}
          &  SVM  & 0.86 $\pm$ 0.0016 & 0.86 $\pm$ 0.0039 & 0.86 $\pm$ 0.0018 & 0.85 $\pm$ 0.0025 & 0.69 $\pm$ 0.0120 & 0.84 $\pm$ 0.0040 & 136.92 $\pm$ 0.2159 \\
          &  NB  & 0.67 $\pm$ 0.0072 & 0.71 $\pm$ 0.0062 & 0.67 $\pm$ 0.0089 & 0.66 $\pm$ 0.0076 & 0.48 $\pm$ 0.0133 & 0.75 $\pm$ 0.0087 & 10.07 $\pm$ 0.2322 \\
          &  MLP  & 0.82 $\pm$ 0.0038 & 0.83 $\pm$ 0.0447 & 0.82 $\pm$ 0.0041 & 0.81 $\pm$ 0.0051 & 0.61 $\pm$ 0.0174 & 0.8 $\pm$ 0.0067 & 69.85 $\pm$ 0.1870 \\
          &  KNN  & 0.81 $\pm$ 0.0098 & 0.81 $\pm$ 0.0055 & 0.81 $\pm$ 0.0107 & 0.8 $\pm$ 0.0084 & 0.61 $\pm$ 0.0246 & 0.8 $\pm$ 0.0135 & 117.44 $\pm$ 2.0245 \\
          &  RF  & 0.86 $\pm$ 0.0056 & 0.85 $\pm$ 0.0071 & 0.86 $\pm$ 0.0061 & 0.84 $\pm$ 0.0065 & 0.68 $\pm$ 0.0122 & 0.84 $\pm$ 0.0045 & 13.02 $\pm$ 0.0847 \\
          &  LR  & 0.87 $\pm$ 0.0041 & 0.87 $\pm$ 0.0037 & 0.87 $\pm$ 0.0045 & 0.85 $\pm$ 0.0055 & 0.69 $\pm$ 0.0296 & 0.84 $\pm$ 0.0148 & 48.76 $\pm$ 0.0924 \\
          &  DT  & 0.86 $\pm$ 0.0085 & 0.85 $\pm$ 0.0083 & 0.86 $\pm$ 0.0092 & 0.85 $\pm$ 0.0087 & 0.68 $\pm$ 0.0208 & 0.83 $\pm$ 0.0123 & 2.45 $\pm$ 0.0709 \\

         \midrule
        \multirow{7}{1.2cm}{PWM2Vec}
          &  SVM  & 0.82 $\pm$ 0.0019 & 0.81 $\pm$ 0.0055 & 0.82 $\pm$ 0.0028 & 0.81 $\pm$ 0.0036 & 0.58 $\pm$ 0.0166 & 0.79 $\pm$ 0.0063 & 17.13 $\pm$ 0.4269 \\
          &  NB  & 0.51 $\pm$ 0.0084 & 0.6 $\pm$ 0.0088 & 0.51 $\pm$ 0.0140 & 0.52 $\pm$ 0.0108 & 0.13 $\pm$ 0.0184 & 0.62 $\pm$ 0.0137 & 0.96 $\pm$ 0.4591 \\
          &  MLP  & 0.8 $\pm$ 0.0044 & 0.78 $\pm$ 0.0631 & 0.8 $\pm$ 0.0065 & 0.77 $\pm$ 0.0073 & 0.47 $\pm$ 0.0239 & 0.73 $\pm$ 0.0106 & 19.02 $\pm$ 0.1650 \\
          &  KNN  & 0.81 $\pm$ 0.0115 & 0.82 $\pm$ 0.0077 & 0.81 $\pm$ 0.0169 & 0.8 $\pm$ 0.0119 & 0.6 $\pm$ 0.0340 & 0.79 $\pm$ 0.0212 & 7.78 $\pm$ 4.0020 \\
          &  RF  & 0.85 $\pm$ 0.0065 & 0.84 $\pm$ 0.0100 & 0.85 $\pm$ 0.0096 & 0.84 $\pm$ 0.0093 & 0.62 $\pm$ 0.0168 & 0.81 $\pm$ 0.0071 & 4.8 $\pm$ 0.1675 \\
          &  LR  & 0.82 $\pm$ 0.0048 & 0.81 $\pm$ 0.0052 & 0.82 $\pm$ 0.0070 & 0.81 $\pm$ 0.0078 & 0.57 $\pm$ 0.0408 & 0.79 $\pm$ 0.0232 & 33.44 $\pm$ 0.1826 \\
          &  DT  & 0.81 $\pm$ 0.0099 & 0.82 $\pm$ 0.0117 & 0.81 $\pm$ 0.0146 & 0.81 $\pm$ 0.0124 & 0.57 $\pm$ 0.0286 & 0.78 $\pm$ 0.0194 & 2.47 $\pm$ 0.1401 \\

         \midrule
        \multirow{7}{1.2cm}{Kernel Approx.}
          &  SVM   & 0.85 $\pm$ 0.0023 & 0.85 $\pm$ 0.0043 & 0.85 $\pm$ 0.0021 & 0.84 $\pm$ 0.0030 & 0.63 $\pm$ 0.0132 & 0.81 $\pm$ 0.0040 & 5.06 $\pm$ 0.2591 \\
          &  NB    & 0.75 $\pm$ 0.0101 & 0.81 $\pm$ 0.0069 & 0.75 $\pm$ 0.0106 & 0.76 $\pm$ 0.0091 & 0.58 $\pm$ 0.0147 & 0.8 $\pm$ 0.0086 & 0.11 $\pm$ 0.2787 \\
          &  MLP   & 0.85 $\pm$ 0.0053 & 0.84 $\pm$ 0.0491 & 0.85 $\pm$ 0.0049 & 0.83 $\pm$ 0.0061 & 0.66 $\pm$ 0.0191 & 0.83 $\pm$ 0.0067 & 15.92 $\pm$ 0.1644 \\
          &  KNN   & 0.82 $\pm$ 0.0137 & 0.82 $\pm$ 0.0060 & 0.82 $\pm$ 0.0128 & 0.82 $\pm$ 0.0100 & 0.62 $\pm$ 0.0271 & 0.79 $\pm$ 0.0133 & 0.29 $\pm$ 2.4294 \\
          &  RF    & 0.85 $\pm$ 0.0078 & 0.85 $\pm$ 0.0078 & 0.85 $\pm$ 0.0073 & 0.84 $\pm$ 0.0078 & 0.66 $\pm$ 0.0134 & 0.82 $\pm$ 0.0044 & 1.49 $\pm$ 0.1017 \\
          &  LR    & 0.85 $\pm$ 0.0057 & 0.84 $\pm$ 0.0040 & 0.85 $\pm$ 0.0053 & 0.83 $\pm$ 0.0066 & 0.6 $\pm$ 0.0325 & 0.81 $\pm$ 0.0146 & 1.76 $\pm$ 0.1108 \\
          &  DT    & 0.83 $\pm$ 0.0119 & 0.83 $\pm$ 0.0091 & 0.83 $\pm$ 0.0111 & 0.82 $\pm$ 0.0104 & 0.63 $\pm$ 0.0228 & 0.81 $\pm$ 0.0122 & 0.25 $\pm$ 0.0850 \\
           \midrule
        \multirow{7}{1.2cm}{OMK}
          &  SVM  & 0.86 $\pm$ 0.0018 & 0.86 $\pm$ 0.0052 & 0.86 $\pm$ 0.0026 & 0.85 $\pm$ 0.0034 & 0.67 $\pm$ 0.0156 & 0.83 $\pm$ 0.0060 & 46.7 $\pm$ 0.4012 \\
          &  NB   & 0.71 $\pm$ 0.0079 & 0.79 $\pm$ 0.0083 & 0.71 $\pm$ 0.0132 & 0.73 $\pm$ 0.0102 & 0.49 $\pm$ 0.0173 & 0.75 $\pm$ 0.0129 & 0.12 $\pm$ 0.4315 \\
          &  MLP  & 0.85 $\pm$ 0.0042 & 0.85 $\pm$ 0.0593 & 0.85 $\pm$ 0.0061 & 0.83 $\pm$ 0.0069 & 0.64 $\pm$ 0.0225 & 0.82 $\pm$ 0.0100 & 30.54 $\pm$ 0.1191 \\
          &  KNN  & 0.83 $\pm$ 0.0108 & 0.85 $\pm$ 0.0073 & 0.83 $\pm$ 0.0159 & 0.83 $\pm$ 0.0112 & 0.64 $\pm$ 0.0319 & 0.82 $\pm$ 0.0199 & 0.27 $\pm$ 3.7619 \\
          &  RF   & 0.86 $\pm$ 0.0061 & 0.86 $\pm$ 0.0094 & 0.86 $\pm$ 0.0090 & 0.84 $\pm$ 0.0087 & 0.65 $\pm$ 0.0158 & 0.82 $\pm$ 0.0066 & 1.43 $\pm$ 0.1574 \\
          &  LR   & 0.87 $\pm$ 0.0045 & 0.87 $\pm$ 0.0049 & 0.87 $\pm$ 0.0066 & 0.86 $\pm$ 0.0073 & 0.69 $\pm$ 0.0383 & 0.84 $\pm$ 0.0218 & 3.1 $\pm$ 0.1716 \\
          &  DT   & 0.86 $\pm$ 0.0093 & 0.86 $\pm$ 0.0110 & 0.86 $\pm$ 0.0137 & 0.85 $\pm$ 0.0117 & 0.68 $\pm$ 0.0269 & 0.83 $\pm$ 0.0182 & 0.19 $\pm$ 0.1317 \\

          \midrule
        \multirow{7}{1.2cm}{IGK}
           &  SVM  & 0.86 $\pm$ 0.0016 & 0.86 $\pm$ 0.0042 & 0.86 $\pm$ 0.0017 & 0.84 $\pm$ 0.0026 & 0.62 $\pm$ 0.0127 & 0.81 $\pm$ 0.0042 & 4.94 $\pm$ 0.2310 \\
          &  NB  & 0.74 $\pm$ 0.0070 & 0.82 $\pm$ 0.0068 & 0.74 $\pm$ 0.0086 & 0.76 $\pm$ 0.0076 & 0.56 $\pm$ 0.0141 & 0.81 $\pm$ 0.0090 &  \textbf{0.08} $\pm$ 0.2484 \\
          &  MLP  & 0.84 $\pm$ 0.0037 & 0.84 $\pm$ 0.0484 & 0.84 $\pm$ 0.0040 & 0.83 $\pm$ 0.0052 & 0.59 $\pm$ 0.0184 & 0.8 $\pm$ 0.0070 & 10.62 $\pm$ 0.1140 \\
          &  KNN  & 0.83 $\pm$ 0.0096 & 0.83 $\pm$ 0.0059 & 0.83 $\pm$ 0.0104 & 0.83 $\pm$ 0.0085 & 0.61 $\pm$ 0.0261 & 0.8 $\pm$ 0.0139 & 0.3 $\pm$ 2.1658 \\
          &  RF  & 0.86 $\pm$ 0.0055 & 0.86 $\pm$ 0.0077 & 0.86 $\pm$ 0.0059 & 0.84 $\pm$ 0.0066 & 0.62 $\pm$ 0.0129 & 0.81 $\pm$ 0.0046 & 1.19 $\pm$ 0.0906 \\
          &  LR  & 0.86 $\pm$ 0.0040 & 0.86 $\pm$ 0.0040 & 0.86 $\pm$ 0.0043 & 0.83 $\pm$ 0.0055 & 0.6 $\pm$ 0.0313 & 0.8 $\pm$ 0.0153 & 1.65 $\pm$ 0.0988 \\
          &  DT  & 0.84 $\pm$ 0.0083 & 0.84 $\pm$ 0.0090 & 0.84 $\pm$ 0.0089 & 0.83 $\pm$ 0.0088 & 0.59 $\pm$ 0.0220 & 0.8 $\pm$ 0.0127 & 0.18 $\pm$ 0.0758 \\
         \midrule
         \multirow{7}{1.2cm}{OMK + IG}
           &  SVM  & 0.87 $\pm$ 0.0020 & 0.87 $\pm$ 0.0038 & 0.87 $\pm$ 0.0019 & 0.85 $\pm$ 0.0027 & 0.69 $\pm$ 0.0118 & 0.84 $\pm$ 0.0036 & 15.09 $\pm$ 0.2306 \\
          &  NB  & 0.76 $\pm$ 0.0090 & 0.84 $\pm$ 0.0061 & 0.76 $\pm$ 0.0095 & 0.77 $\pm$ 0.0081 & 0.6 $\pm$ 0.0130 & 0.83 $\pm$ 0.0077 & 0.1 $\pm$ 0.2480 \\
          &  MLP  & 0.86 $\pm$ 0.0047 & 0.85 $\pm$ 0.0437 & 0.86 $\pm$ 0.0044 & 0.85 $\pm$ 0.0055 & 0.66 $\pm$ 0.0170 & 0.83 $\pm$ 0.0059 & 18.67 $\pm$ 0.1133 \\
          &  KNN  & 0.85 $\pm$ 0.0122 & 0.85 $\pm$ 0.0054 & 0.85 $\pm$ 0.0114 & 0.85 $\pm$ 0.0089 & 0.65 $\pm$ 0.0241 & 0.81 $\pm$ 0.0119 & 0.3 $\pm$ 2.1622 \\
          &  RF  &  \textbf{0.88}  $\pm$ 0.0070 &  \textbf{0.88}  $\pm$ 0.0069 &  \textbf{0.88}  $\pm$ 0.0065 &  \textbf{0.87}  $\pm$ 0.0069 &  \textbf{0.70}  $\pm$ 0.0119 &  \textbf{0.85}  $\pm$ 0.0040 & 1.29 $\pm$ 0.0905 \\
          &  LR  & 0.87 $\pm$ 0.0051 & 0.87 $\pm$ 0.0036 & 0.87 $\pm$ 0.0048 & 0.86 $\pm$ 0.0058 & 0.68 $\pm$ 0.0290 & 0.84 $\pm$ 0.0130 & 2.37 $\pm$ 0.0986 \\
          &  DT  & 0.85 $\pm$ 0.0106 & 0.85 $\pm$ 0.0081 & 0.85 $\pm$ 0.0099 & 0.84 $\pm$ 0.0093 & 0.66 $\pm$ 0.0203 & 0.83 $\pm$ 0.0108 & 0.21 $\pm$ 0.0757 \\

        \bottomrule
    \end{tabular}
    }
    \caption{Average $\pm$ standard deviation classification results for GISAID-2 dataset. Best values are shown in bold.}
    \label{tbl_classification_gisaid2}
\end{table*}

\subsection{Clustering Results}
To evaluate the performance of different methods in terms of clustering, we report the quality of clustering using different evaluation metrics. The clustering results for GISAID-1 and GISAID-2 datasets are given in Table~\ref{tbl_clust_1} and Table~\ref{tbl_clust_2}, respectively. For the silhouette coefficient, OMK method performs better than the other methods for both datasets. In terms of Calinski-Harabasz score, IGK performs better than the baselines for both datasets.
However, the OHE outperforms all methods in terms of Davies-Bouldin score. However, one problem with OHE method is its runtime complexity due to high dimensionality of the vectors.
In terms of runtime, OMK + IG performs better than other methods in case of GISAID-1 dataset, while OMK outperforms the other methods in case of GISAID-2 dataset. From the reported clustering results, we can conclude that there is no single method that outperforms all other approaches for all evaluation metrics (as can be seen from classification results). However, kernel-based methods appear to be performing better overall for both datasets (similar behavior is observed from classification results).

\begin{table}[h!]
  \centering
  \resizebox{0.49\textwidth}{!}{
  \begin{tabular}{p{2cm}p{1.4cm}p{2.1cm}p{1.9cm}p{0.9cm}}
    \toprule
    \multirow{2}{*}{Methods} & Silhouette Coefficient & Calinski-Harabasz Score & Davies-Bouldin Score & Runtime (Sec.) \\
    \midrule	\midrule	
    OHE & 0.856 & 32376.919 & \textbf{0.250} & 24.77 \\
    Spike2Vec & 0.834 & 22794.361 & 0.467 & 11.31 \\
    PWM2Vec & 0.477 & 1762.983 & 1.007 & 1.45 \\
    Kernel Approx. & 0.851 & 24619.646 & 0.423 &  0.078 \\
    OMK &  \textbf{0.858} & 22083.103 & 0.456 & 0.080 \\
    IGK &  0.717 & \textbf{37924.721} & 0.489 & 0.093 \\
    OMK + IG &  0.672 & 14459.024 & 0.578 & \textbf{0.073} \\
    \bottomrule
  \end{tabular}
  }
  \caption{Clustering performance comparison using different metrics for k-means on GISAID-1 dataset. Best values are shown in bold.}
  \label{tbl_clust_1}
\end{table}

\begin{table}[h!]
  \centering
  \resizebox{0.49\textwidth}{!}{
  \begin{tabular}{p{2cm}p{1.4cm}p{2.1cm}p{1.9cm}p{0.9cm}}
    \toprule
    \multirow{2}{*}{Methods} & Silhouette Coefficient & Calinski-Harabasz Score & Davies-Bouldin Score & Runtime (Sec.) \\
    \midrule	\midrule	
    OHE & 0.862 & 34223.922 & \textbf{0.239} & 20.48 \\
    Spike2Vec & 0.840 & 30878.005 & 0.465 & 11.32 \\
    PWM2Vec & 0.487 & 2061.861 & 1.033 & 1.27 \\
    Kernel Approx. & 0.863 & 34296.208 & 0.425 & 0.096 \\
    OMK &  \textbf{0.864} & 33966.355 & 0.425 & \textbf{0.058} \\
    IGK &  0.714 & \textbf{35821.496} & 0.502 & 0.088 \\
    OMK + IG &  0.645 & 17086.919 & 0.562 & 0.075 \\
    \bottomrule
  \end{tabular}
  }
  \caption{Clustering performance comparison using different metrics for k-means on GISAID-2 dataset. Best values are shown in bold.}
  \label{tbl_clust_2}
\end{table}

\subsection{Kernel Computation Runtime}
We report the kernel computation runtime for Approximate kernel, IGK, OMK, and OMK + IG in Table~\ref{tbl_kernel_runtime} (for GISAID-1 dataset). We can observe that since IGK contains the least number of amino acids in each sequence, the kernel computation time for this method is the minimum. However, OMK method takes the maximum ($2163$ sec.) to compute the kernel matrix. Since both GISAID-1 and GISAID-2 datasets contain the same number of sequences, the kernel computation time for both datasets will be similar.

\begin{table}[h!]
    \centering
    \begin{tabular}{ccc}
    \toprule
    Method & Runtime (sec.) & \# of Amino Acids \\
    \midrule \midrule
     OMK & 2163.02  & 3798 \\
     OMK + IG & 1818.05  & 2184 \\
     Kernel Approx.  &  1510.07  & 1274  \\
     IGK    & 1048.03 & 243 \\
     \bottomrule
    \end{tabular}
    \caption{Kernel computation runtime for different methods. Note that since both GISAID-1 and GISAID-2 dataset have $7000$ sequences each, the kernel computation runtime for both datasets will be similar. The last column shows the number of amino acids in the input data for kernel matrices.}
    \label{tbl_kernel_runtime}
\end{table}

\section{Conclusion}\label{sec_conclusion}

The COVID-19 outbreak induced by SARS-CoV-2 captured the scientific community's attention across the world. The current research on SARS-CoV-2 focused on understanding the transmission pattern of the virus, identifying new variants, improving public health, and developing start-of-art vaccine and treatment options. Computational biology played a significant role in this scientific journey of comprehensive understanding of the COVID-19 pandemic and management. Especially in the processing of the high-throughput sequencing data, there is an unmet need to classify the genomic data accurately.

Here, we propose three different settings to efficiently perform different machine learning tasks such as classification and clustering on SARS-CoV-2 variants
using spike sequences. Results show that the minimizer plus information gain-based method outperforms the existing baseline and state-of-the-art methods in terms of predictive performance.  
In the future, we will work towards detecting new (unknown) variants (such as Omicron) based on whole genome sequences. We will collect more data in the future to test the scalability of the proposed model.
Another exciting future work is considering other attributes like
countries, cities, and dates to design richer feature vector
representations for spike sequences.

\section*{Acknowledgements}

The authors would like to acknowledge funding from an MBD fellowhip to Sarwan Ali and Bikram Sahoo, and Georgia State University Computer Science Startup Grant to Murray Patterson.

\bibliographystyle{IEEEtran}
\bibliography{bibliography}

\end{document}